\documentclass[5p,12pt]{elsarticle}
\usepackage{amsmath}
\usepackage{amsfonts}
\usepackage{amssymb}
\usepackage{graphicx}
\usepackage{verbatim}
\usepackage{xcolor}
\usepackage{bm}
\usepackage[ruled, lined, linesnumbered, commentsnumbered]{algorithm2e} 
\usepackage{subcaption}
\usepackage{multirow}
\usepackage{pdflscape}
\immediate\write18{texcount -tex -sum  \jobname.tex > \jobname.wordcount.tex}
\usepackage{nomencl}
\makenomenclature
\usepackage{tikz}
\usepackage[acronym,nonumberlist,nopostdot]{glossaries}

\providecommand{\keywords}[1]
{
  \small	
  \textbf{\textit{}} #1
}

\title{
Efficient Training of Learning-Based Thermal Power Flow for 4\Th Generation District Heating Grids
}
\author{Andreas Bott$^{1}$, Mario Beykirch$^{1}$, Florian Steinke$^{1}$ \\
        \small $^{1}$ Energy Information Networks \& Systems, Technical University of Darmstadt, Landgraf-Georg-Str. 4, 64283 Darmstadt, Germany\\
}
\date{} % Comment this line to show today's date

\newcommand{\mycomment}[1]{}

\newcommand{\Th}{\textsuperscript{th} }
\newcommand{\Rd}{\textsuperscript{rd} }

\newcommand{\abs}[1]{\left| #1 \right|}
  \newcommand{\Norm}[1]{\|#1\|_2}
  \DeclareMathOperator*{\Margmin}{arg\,min}
  \newcommand{\argmin}[1]{\Margmin_{#1}}
  \newcommand{\Sum}[2]{\Sigma_{#1}^{#2}}
  \DeclareMathOperator*{\Supp}{supp}
  \newcommand{\supp}[1]{\Supp(#1)}

  \DeclareMathOperator*{\MVar}{var}
  \newcommand{\Var}[1]{\MVar(#1)}
  \newcommand{\C}{^\circ C}

  \newcommand{\fig}[1]{Figure~\ref{fig:#1}}
  \newcommand{\tab}[1]{Table~\ref{tab:#1}}
  \newcommand{\apdx}[1]{\ref{app:#1}}
  \newcommand{\alg}[1]{Algorithm~\ref{alg:#1}}

\usepackage{etoolbox}
\renewcommand\nomgroup[1]{%
  \item[\bfseries
  \ifstrequal{#1}{A}{General Abbreviations}{%
  \ifstrequal{#1}{B}{Grid Parametrisation}{%  
  \ifstrequal{#1}{H}{State Variables}{%
  \ifstrequal{#1}{S}{Classic Solution Algorithm}{%
  \ifstrequal{#1}{T}{Stochastic Variables}{%
  \ifstrequal{#1}{Z}{Leraning Based Models}{%
  }}}}}}%
]}
  \newcommand{\Grid}{G}
  \nomenclature[B, 110]{$\Grid$}{Graph describing the heating grid $G = (\Nodes ,\Edges)$}
  \newcommand{\Edges}{E}
  \nomenclature[B, 112]{$\Edges$}{Edges of the graph $\Grid$; $\Edges = \EDGp \cup \EDGa \cup \EDGsl$}
  \newcommand{\EDGp}{\Edges^{passive}}
  \nomenclature[B, 121]{$\EDGp$}{Pipes in the network}
  \newcommand{\EDGa}{\Edges^{active}}
  \nomenclature[B, 122]{$\EDGa$}{Heat supplies or demands with external power target}
  \newcommand{\EDGsl}{\Edges^{slack}}
  \nomenclature[B, 123]{$\EDGsl$}{Heat supply balancing total power supply and consumption}
  \newcommand{\Nodes}{V}
  \nomenclature[B, 111]{$\Nodes$}{Nodes of the graph $\Grid$}
  \newcommand{\Neighbours}[1]{N_{#1}}
  \nomenclature[B, 125]{$\Neighbours{i}$}{Set of nodes neighbouring node $i$}

  \newcommand{\Ladder}[2]{ladder^{#1}_{#2}}
  \newcommand{\Cycle}[2]{cycle^{#1}_{#2}}
  \nomenclature[B, 200]{$\Ladder{na}{sp}$}{Grid with line structure, $na$ active components and supplies at   positions $sp$}
  \nomenclature[B, 201]{$\Cycle{na}{sp}$}{Grid with looping structure, $na$ active components   and supplies at positions $sp$}
  
  \nomenclature[B, 151]{$a_{ij}$}{Heat loss coefficient of pipe ${(i,j) \in \EDGp}$ }
  \newcommand{\pressurdrop}[1]{k_{#1}}
  \nomenclature[B, 152]{$\pressurdrop{i,j}$}{Pressure loss factor of pipe $(i,j) \in \EDGp$}
  \newcommand{\tempa}{\temp{a}}
  \nomenclature[B, 153]{$\tempa$}{Ambient temperature}
  \newcommand{\cp}{c_{p}}
  \nomenclature[B, 154]{$\cp$}{Specific heat capacity of water}

  \renewcommand{\temp}[1]{T_{#1}}
  \newcommand{\vtemp}{\mathbf{\temp{}}}
  \nomenclature[H, 110]{$\temp{i}$}{Temperature at node ${i \in \Nodes}$}
  \nomenclature[H, 111]{$\vtemp$}{Vector containing all temperatures \\ ${\vtemp = [\temp{i} \forall i \in \Nodes]}$}
  \newcommand{\pr}[1]{p_{#1}}
  \newcommand{\vpr}{\mathbf{\pr{}}}
  \nomenclature[H, 112]{$\pr{i}$}{Pressure at node ${i \in \Nodes}$}
  \nomenclature[H, 113]{$\vpr$}{Vector containing all pressures ${\vpr = [\pr{i} \forall i \in \Nodes]}$}
  \newcommand{\mf}[1]{\dot{m}_{#1}}
  \newcommand{\vmf}{\mathbf{\dot{m}}}
  \nomenclature[H, 114]{$\mf{ij}$}{Mass flow through edge $(i,j)$}
  \nomenclature[H, 115]{$\vmf$}{Vector containing all mass flows \\ ${\vmf = [\mf{ij} \forall (i,j) \in \Edges]}$}
  \newcommand{\tempend}[1]{\temp{#1}^{end}}
  \newcommand{\vtempend}{\vtemp^{end}}
  \nomenclature[H, 117]{$\tempend{ij}$}{Temperature at the outlet of edge $(i,j)$}
  \nomenclature[H, 118]{$\vtempend$}{Vector containing all end of line temperatures ${\vtempend =   [\tempend{ij} \forall (i,j) \in \Edges]}$}
  \newcommand{\tempstart}[1]{\temp{#1}^{start}}
  \nomenclature[H, 119]{$\tempstart{ij}$}{Temperature at the inlet of edge $(i,j)$}
  \newcommand{\power}[1]{\dot{q}_{#1}}
  \newcommand{\vpower}{\mathbf{\dot{q}}}
  \nomenclature[H, 120]{$\power{ij}$}{Heat power exchanged at edge $(i,j) \in \EDGa$}
  \nomenclature[H, 121]{$\vpower$}{Vector containing all power values \\${\vpower = [\power{ij} \forall (i,j) \in \EDGa]}$}
  \newcommand{\tfi}[1]{\temp{#1}^{fi}}
  \newcommand{\vtfi}{\vtemp^{fi}}
  \nomenclature[H, 122]{$\tfi{ij}$}{Feed in temperature of active edge \\$(i,j) \in \EDGa \cup \EDGsl$}
  \nomenclature[H, 123]{$\vtfi$}{Vector containing all feed-in temperatures \\${\vtfi = [\tfi{ij} \forall (i,j) \in \EDGa \cup \EDGsl]}$}
  \newcommand{\prset}[1]{\pr{#1}^{set}}
  \nomenclature[H, 124]{$\prset{i}$}{Pressure setpoint for node $i$}
  \newcommand{\state}{\mathbf{x}}
  \nomenclature[H, 150]{$\state$}{Grid state ${\state = [\vtemp, \vmf, \vpr, \vtempend]}$}
  \newcommand{\statequations}[1]{\mathbf{e}\left(#1\right)}
  \nomenclature[H, 150]{$\statequations{\cdot}$}{State equations $\statequations{\state, \vpower, \vtfi} = 0$}
  \newcommand{\modelDS}[1]{\mathbf{h}\left(#1\right)}
  \nomenclature[H, 151]{$\modelDS{\cdot}$}{State mapping $\modelDS{\vpower, \vtfi} = \state$}
  
  \newcommand{\SELoss}{\Psi}
  \nomenclature[S, 160]{$\SELoss$}{Squared violation of state equations}

  \newcommand{\vtinit}{\vtemp^{init}}
  \nomenclature[S, 171]{$\vtinit$}{Initial temperature assumptions}
  \newcommand{\vmfactive}{\mathbf{\dot{m}^{a}}}
  \nomenclature[H, 116]{$\vmfactive$}{Mass flow through active edges \\${\vmfactive = [\mf{ij} \forall (i,j) \in \EDGa]}$}
  \newcommand{\precisionDA}{\epsilon_{DA}}
  \nomenclature[S, 175]{$\precisionDA$}{Convergence precision for the decomposed algorithm}

  \newcommand{\precisionNR}{\epsilon_{NR}}
  \nomenclature[S, 180]{$\precisionNR$}{Convergence precision for the NR algorithm}
  \newcommand{\stateinit}{\state_{init}}
  \nomenclature[S, 181]{$\stateinit$}{Initial state assumption}
  \newcommand{\Jacobian}{\mathbf{J}}
  \nomenclature[S, 183]{$\Jacobian$}{Jacobian matrix}
  \newcommand{\descdir}{\state^{des}}
  \nomenclature[S, 184]{$\descdir$}{Descent direction during NR algorithm}
  \newcommand{\stepsize}{\alpha}
  \nomenclature[S, 185]{$\stepsize$}{Stepsize for the NR algorithm}
  \newcommand{\decparam}{\gamma}
  \nomenclature[S, 186]{$\decparam$}{Control parameter for the backtracking used during the NR algorithm}

  \newcommand{\modelWeight}{\bm{\theta}}
  \newcommand{\modelWeightOpt}{\modelWeight^{*}}
  \newcommand{\modelNN}[1]{\mathbf{h}_{\modelWeight}\left({#1}\right)}
  \nomenclature[Z, 152]{$\modelNN{\cdot}$}{Approximation of the state mapping \\ $\modelNN{\vpower, \vtfi} \approx \modelDS{\vpower, \vtfi}$}
  \nomenclature[Z, 153]{$\modelWeight$}{Weights of $\modelNN{\cdot}$}
  \nomenclature[Z, 154]{$\modelWeightOpt$}{Optimal Value for $\modelWeight$}
  \newcommand{\statepred}{\hat{\state}}
  \newcommand{\vpowerpred}{\hat{\dot{\bm{q}}}}
  \nomenclature[Z, 155]{$\statepred$}{State prediction by the learned model}
  \nomenclature[Z, 156]{$\vpowerpred$}{Predicted heating power calculated based on $\statepred$}

  \newcommand{\Loss}[1]{L(#1)}
  \nomenclature[Z, 157]{$\Loss{\cdot}$}{Loss function of the learning-based models}
  \newcommand{\LossPA}[1]{L_{PA}(#1)}
  \nomenclature[Z, 158]{$\LossPA{\cdot}$}{Physics aware loss function}
  \newcommand{\LossW}[1]{L_{\omega'}(#1)}
  \nomenclature[Z, 159]{$\LossW{\cdot}$}{Weighted L2-loss function}
  \newcommand{\weightLoss}[1]{\omega'_{#1}}
  \nomenclature[Z, 160]{$\weightLoss{\kappa}$}{Weighting factor for different dimensions in the loss function}
  \newcommand{\Ntrain}{N}
  \nomenclature[Z, 161]{$\Ntrain$}{Number of training samples}
  
  \newcommand{\weightSample}[1]{\omega_{#1}}
  \nomenclature[T, 170]{$\weightSample{i}$}{Weight of sample $i$ to compensate for importance sampling}
  \newcommand{\ldem}[1]{d_{#1}}
  \newcommand{\demcor}[1]{\rho_{#1}}
  \newcommand{\corrfactor}{\lambda_{c}}
  \nomenclature[T, 175]{$\ldem{}$}{Distance between two demands}
  \nomenclature[T, 176]{$\demcor{i,j}$}{Cross-correlation coefficient between power values of demands $i$ and $j$}
  \nomenclature[T, 177]{$\corrfactor$}{Correlation-strength-factor}
  \newcommand{\weightsampleNorm}[1]{\bar{\weightSample{}}_{#1}}
  \nomenclature[T, 171]{$\weightsampleNorm{i}$}{Normalised weight of sample $i$}
  
  \newcommand{\vdist}[1]{\mathbf{P}(#1)}
  \nomenclature[T, 098]{$\vdist{\mathbf{a}}$}{Distribution over random vector $\mathbf{a}$}
  \newcommand{\prob}[1]{f_{\mathbf{P}}(#1)}
  \nomenclature[T, 099]{$\prob{\mathbf{a}}$}{Probability density function of random vector $\mathbf{a}$}
  
  \newcommand{\ZTN}[1]{\bm{N}^{0}\left(#1\right)}
  \newcommand{\vpowermean}{\bm{\mu}_{\dot{q}}}
  \newcommand{\vpowercov}{\bm{\Sigma}_{\dot{q}}}
  \newcommand{\vpowerstd}{\bm{\sigma}_{\dot{q}}}
  \newcommand{\vpowercor}{\bm{\rho}_{\dot{q}}}
  \nomenclature[T, 100]{$\ZTN{\bm{\mu}, \bm{\Sigma}}$}{Zero truncated normal distribution with mean   $\bm{\mu}$ and covariance matrix $\bm{\Sigma}$}
  \nomenclature[T, 101]{$\vpowermean$}{Mean power values for active edges}
  \nomenclature[T, 102]{$\vpowerstd$}{Vector containing the standard deviations for power values}
  \nomenclature[T, 103]{$\vpowercov$}{Covariance matrix for power values}
  \nomenclature[T, 104]{$\vpowercor$}{Normalised correlation-matrix for power values}
  \newcommand{\vmfactivestd}{\bm{\sigma}_{\dot{m}}}
  \newcommand{\vmfactivemean}{\bm{\mu}_{\dot{m}}}
  \newcommand{\vmfactivecov}{\bm{\Sigma}_{\dot{m}}}
  \newcommand{\vmfactivecor}{\bm{\rho}_{\dot{m}}}
  \nomenclature[T, 201]{$\vmfactivemean$}{Mean mass flow values for active edges}
  \nomenclature[T, 202]{$\vmfactivestd$}{Vector containing the standard deviations for active edges' mass flows}
  \nomenclature[T, 203]{$\vmfactivecov$}{Covariance matrix for active edges' mass flows}
  \nomenclature[T, 204]{$\vmfactivecor$}{Normalised correlation-matrix for active edges' mass flows}
  \newcommand{\vtfimin}{\vtfi_{min}}
  \newcommand{\vtfimax}{\vtfi_{max}}
  \nomenclature[T, 303]{$\vtfimin$}{Vector containing the minimal feed-in temperatures}
  \nomenclature[T, 304]{$\vtfimax$}{Vector containing the maximum feed-in temperatures}  
  \newcommand{\effSR}{ESR}
  \nomenclature[T, 372]{$\effSR$}{Effective sample rate of importance sampling algorithm}
  
  \newcommand{\LossSE}{\Psi}
  \newcommand{\stateequationsPas}[1]{\bar{\mathbf{e}}\left(#1\right)}

\newacronym[description={Newton-Raphson algorithm}]{NR}{NR}{Newton-Raphson algorithm}
\newacronym[description={Decomposed algorithm}]{DC}{DC}{decomposed algorithm}
\newacronym[description={Importance-sampling like algorithm}]{IS}{IS}{importance-sampling like algorithm}
\newacronym[description={Deep Neural Network}]{DNN}{DNN}{Deep Neural Network}
\newacronym[description={Rectified Linear Unit}]{ReLU}{ReLU}{rectified linear unit}
\newacronym[description={District Heating}]{DH}{DH}{District Heating}
\newacronym[description={Machine Learning}]{ML}{ML}{Machine Learning}
\newacronym[description={Mean Absolute Error}]{MAE}{MAE}{mean absolute error}
\newacronym[description={Root Mean Squared Error}]{RMSE}{RMSE}{root mean squared error}
\newacronym[description={Thermal power flow}]{TPF}{TPF}{thermal power flow}

\makeglossaries

\begin{document}
% \begin{abstract}
\begin{abstract}
    % 4\Th generation district heating grids are a key tool to decarbonize the heat sector.
    % Multiple decentral heat sources and meshed grid structures render thermal power flow in these grids an important task for many control purposes.   
    Thermal power flow (TPF) is an important task for various control purposes in 4 Th generation district heating grids with multiple decentral heat sources and meshed grid structures. 
    Computing the TPF, i.e., determining the grid state consisting of temperatures, pressures, and mass flows for given supply and demand values, is classically done by solving the nonlinear heat grid equations, but can be sped up by orders of magnitude using learned models such as neural networks. 
    We propose a novel, efficient scheme to generate a sufficiently large training data set covering relevant supply and demand values. 
    Instead of sampling supply and demand values, our approach generates training examples from a proxy distribution over generator and consumer mass flows, omitting the iterations needed for solving the heat grid equations. 
    The exact, but slightly different, training examples can be weighted to represent the original training distribution. 
    We show with simulations for typical grid structures that the new approach can reduce training set generation times by two orders of magnitude compared to sampling supply and demand values directly, without loss of relevance for the training samples. 
    Moreover, learning TPF with a training data set is shown to outperform sample-free, physics-aware training approaches significantly.

\end{abstract}

\maketitle

\hspace{10pt}
\printnomenclature
\printglossaries
%TC:ignore

\section{Introduction} \label{sec:introduction}
Heating and cooling accounts for half of Europe's gross final energy consumption \cite{noauthor_decarbonising_nodate} and decarbonising this sector is thus essential for reaching CO\textsc{2} emission reduction goals \cite{noauthor_proposal_2021}.
\gls{DH} grids can support this effort, which is why 
%by interconnecting renewable heat sources, heat storage and consumers.
expanding grid-based heat supplies is part of various, current legislation activities \cite{noauthor_proposal_2021, klimaschutz_entwurf_nodate}. 
To reduce carbon emissions, the heating sources of \gls{DH} grids must diversify and \gls{DH} grids have to be operated more flexibly \cite{paardekooper_heat_2022}.
This is often associated with a transition from so-called 3\Rd generation to 4\Th generation \gls{DH} grids \cite{lund20144th}.
A key feature of 4\Th generation \gls{DH} grids are low supply temperatures, which reduce grid losses compared to traditional high-temperature grids and allows for the integration of several low-carbon, low-temperature heat sources such as waste heat and heat pumps \cite{mathiesen2015ida, prina2016smart}. 
Additionally, such grids can utilise heat storages and integrate flexible consumers reacting to fluctuating renewable heat availability \cite{van2017dynamic}. 
Loop-based grid layouts are common for 4\Th generation \gls{DH} since they can improve the distribution of heat flows and thereby allow for the easier integration of decentral heat sources \cite{Flexynets2016}.

Due to various, decentralised heat supplies and loops in the network, mass and power flows in 4\Th generation \gls{DH} grids can change significantly over time and even in direction. 
This is challenging for grid operators, who have to guarantee sufficiently high supply temperatures and sufficient pressure drops at each consumer. Advanced grid transparency and control methods are therefore key enablers for 4\Th generation \gls{DH} grids \cite{novitsky2020smarter}.
The calculation of the \gls{TPF}, i.e., determining the grid state consisting of temperatures, pressures and mass flows depending on the heat generation and consumption values, is an integral part of many such methods \cite{bott2023deep,boussaid2023evaluation}. 

\bigskip
Classic \gls{TPF} computations iteratively solve a nonlinear system of equations using the \gls{DC} or a non-linear solver such as the \gls{NR} \cite{sarbu2019review, olsthoorn2016integration}. 
This is often computationally burdensome.
Algorithms exploiting the grid topology can improve run-times by low factors, but remain iterative \cite{LIU20161238, tol2021development}.
%however, to our knowledge, no prior work focused on the task of training data generation, in which the input values are arbitrary, as long as they represent the relevant input space. 
%By abusing this randomness, the computational burden to generate a large number of samples can be reduced significantly. 
In contrast, learning-based approaches without iterations can be many orders of magnitude faster compared to the classic solvers. 
In \cite{bott2023deep}, the authors report calculation times in the order of seconds for a classic solver and microseconds for a neural network model. 
This speed-up can be used, for example, for near real-time, distribution-free, state estimation via Markov Chain Monte Carlo sampling.
The authors of \cite{boussaid2023evaluation} aim in a similar direction, as they propose to use graph neural networks to approximate the quasi-dynamic heat grid equations for reducing the computational burden for optimal control schemes. 
Despite this potential, learning-based approaches to computing the \gls{TPF} are still scarcely used. 
In 2021, the review \cite{mbiydzenyuy_opportunities_2021} found that the most common applications of \gls{ML} methods in the area of \gls{DH} grids are demand prediction, which remains prominent, e.g., \cite{leiprecht_comprehensive_2021, sakkas_thermal_2022, ogliari_machine_2022, habib_hybrid_2023, runge_comparison_2023}, leakage detection, and the modelling and control of complex components such as heat pumps \cite{yabanova_development_2013} or multi-source heat supplies \cite{en13246714}.

\bigskip
In this work, we focus on reducing the computational cost of the model-building process for learning-based approaches to \gls{TPF} calculations.
The process consists of generating synthetic training data examples for relevant grid situations and then actually training the model.
Especially the training data generation step can be computationally prohibitive for larger networks if the default approach is taken to sample typical heat consumption and generation values as inputs for the power flow computations and to determine the corresponding output states for each sample via the classic iterative solvers.

\begin{figure}
    \centering
    \includegraphics[width=0.8\linewidth]{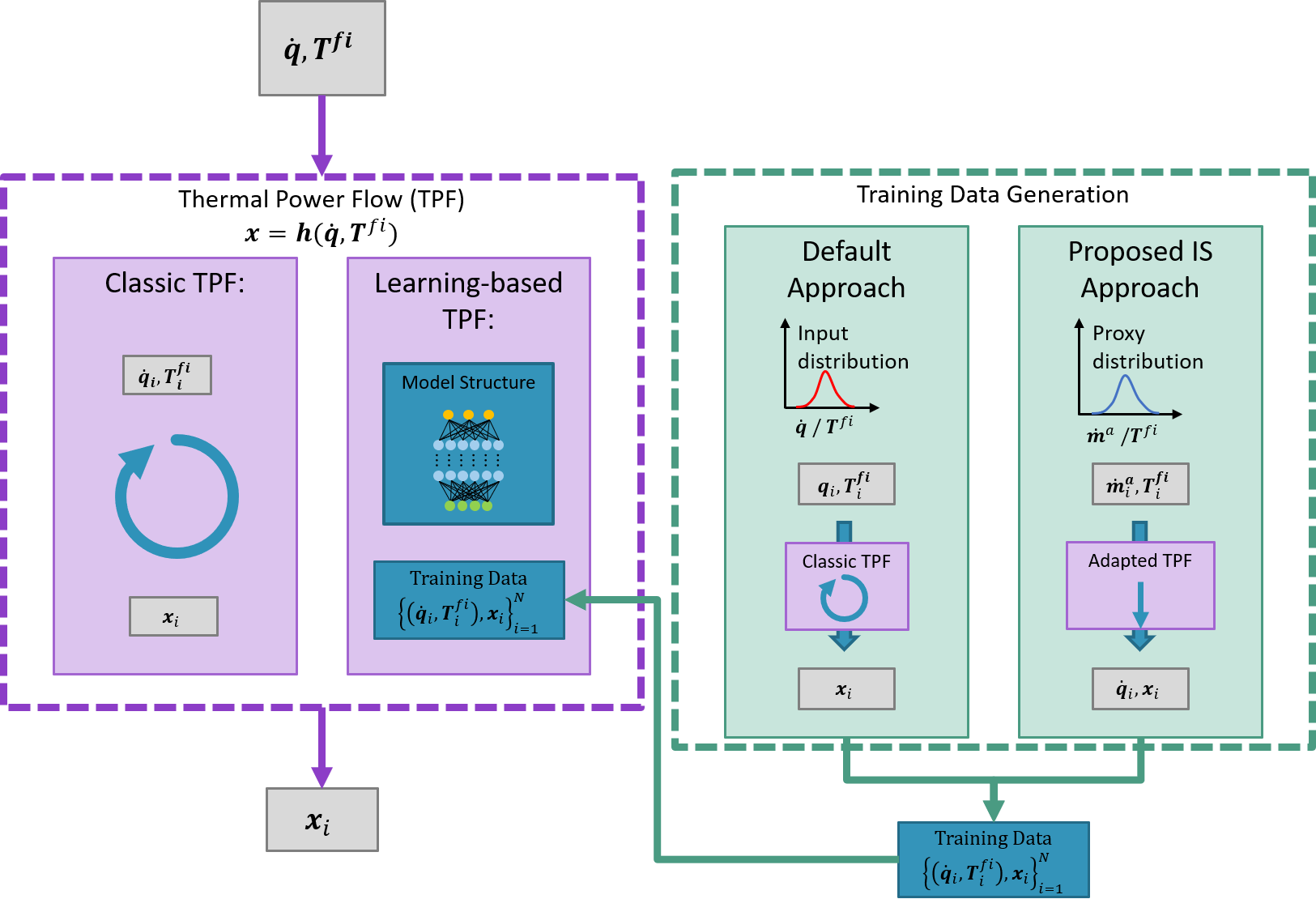}
    \caption{Learning-based approaches can be used to approximate \gls{TPF} computations (purple boxes). 
    The training of such models requires a large number of training samples (blue box). 
    A default way to generate such training samples is to define a distribution over the relevant input values and solve the \gls{TPF} for each sample (left green box). 
    However, classic \gls{TPF} computation methods, such as the \gls{DC} and \gls{NR} algorithms, have an iterative structure, which leads to high computational costs. 
    We propose a novel, importance-sampling-based algorithm to generate training samples (right green box). 
    By defining a suitable proxy distribution over the space of mass flows at heat producers and consumers, we are able to generate each training data point non-iteratively in a single pass, thus drastically reducing computation times for the model-building process.}
    \label{fig:Fig1}
\end{figure}

Instead, we propose a novel, importance-sampling-based algorithm to efficiently determine a suitable training data set in a non-iterative way, as sketched in \fig{Fig1}.
To this end, we first determine a distribution over the mass flows at heat supply and consumer nodes approximating the desired training distribution of heat powers at these sites. 
The grid state and power values for samples from this distribution can then be determined non-iteratively and potential differences between the two distributions can be compensated by weighting the resulting training data set.

In our experiment section, we examine different grid layouts featuring typical characteristics of 4\Th generation \gls{DH} networks. 
We show, that our approach reduces computation times for training sample generation by up to two orders of magnitude and scales favourably with increasing grid complexity compared to the benchmark approach. 
Using a \gls{DNN} as a learning-based model, we additionally compare the training with pre-computed training data against a sample-free physics-aware training procedure, obtaining significantly better error scores with the former approach.
This proves the relevance of our novel, efficient training data generation scheme.
Note that we use a \gls{DNN} for our experiments to illustrate the potential of learning-based approaches. However, the same workflow could be applied to other \gls{ML}-based approaches, as they require similar training data. 

The remainder of this paper is structured as follows: 
The underlying steady-state heat grid model equations are described in section~\ref{sec:model}. 
We then discuss classic algorithms and learning-based approaches to solving the \gls{TPF} problem in section~\ref{subsec:classical_solver} and section~\ref{subsec:LBmodels}, respectively. 
Our main contribution is outlined in section~\ref{sec:IS}, where we present a novel algorithm to generate training data for learning-based approaches.
We conduct numerical experiments in section~\ref{sec:exp} and conclude in section~\ref{sec:concl}.

\section{Heat Grid Equations} \label{sec:model}
\begin{figure}[t]
    \centering
    \begin{subfigure}[b]{\linewidth}
        \centering
        \includegraphics[width=\linewidth]{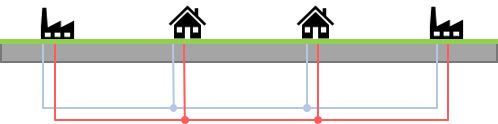} 
        \caption{Topology of a \gls{DH} grid}
    \end{subfigure}
    \hfill
    \\
    \begin{subfigure}[b]{\linewidth}
        \centering
        \scalebox{0.8}{% colors:  	white, black, red, green, blue, cyan, magenta, yellow 
\resizebox{\linewidth}{!}{%
    \begin{tikzpicture}
        % picture parameter: 
        \newcommand{\X}{5}
        \newcommand{\Y}{2}
        \newcommand{\dt}{0.5}
        \newcommand{\cpassive}{blue}
        \newcommand{\cactive}{green}
        \newcommand{\cslack}{red}
        
        % Supply side: 
        \node at (0*\X, 1*\Y) [circle, draw] (01) {$1$};
        \node at (1*\X, 1*\Y) [circle, draw] (02) {$2$};
        \node at (2*\X, 1*\Y) [circle, draw] (03) {$3$};
        \node at (3*\X, 1*\Y) [circle, draw] (04) {$4$};
        % second row:
        \node at (0*\X, 0*\Y) [circle, draw] (11) {$5$};
        \node at (1*\X, 0*\Y) [circle, draw] (12) {$6$};
        \node at (2*\X, 0*\Y) [circle, draw] (13) {$7$};
        \node at (3*\X, 0*\Y) [circle, draw] (14) {$8$};
        % return side: 
        \node at (0*\X, -1*\Y) [circle, draw] (21) {$9 $};
        \node at (1*\X, -1*\Y) [circle, draw] (22) {$10$};
        \node at (2*\X, -1*\Y) [circle, draw] (23) {$11$};
        \node at (3*\X, -1*\Y) [circle, draw] (24) {$12$};
        % second row: 
        \node at (0*\X, -2*\Y) [circle, draw] (31) {$13$};
        \node at (1*\X, -2*\Y) [circle, draw] (32) {$14$};
        \node at (2*\X, -2*\Y) [circle, draw] (33) {$15$};
        \node at (3*\X, -2*\Y) [circle, draw] (34) {$16$};        
        % supply pipes: 
        \draw[ultra thick, color=\cpassive, -latex] (01) -- (02);
        \draw[ultra thick, color=\cpassive, -latex] (02) -- (03);
        \draw[ultra thick, color=\cpassive, -latex] (03) -- (04);
        \draw[ultra thick, color=\cpassive, -latex] (11) -- (01);
        \draw[ultra thick, color=\cpassive, -latex] (02) -- (12);
        \draw[ultra thick, color=\cpassive, -latex] (03) -- (13);
        \draw[ultra thick, color=\cpassive, -latex] (14) -- (04);
        \node at (0.5*\X, 1*\Y+\dt)  [] () {$(1,2)$};
        \node at (1.5*\X, 1*\Y+\dt)  [] () {$(2,3)$};
        \node at (2.5*\X, 1*\Y+\dt) [] () {$(3,4)$};
        \node at (-\dt, 0.5*\Y) [rotate=90] () {$(5,1)$};
        \node at (\X-\dt, 0.5*\Y)  [rotate=90] () {$(2,6)$};
        \node at (2*\X-\dt, 0.5*\Y)  [rotate=90] () {$(3,7)$}; 
        \node at (3*\X-\dt, 0.5*\Y) [rotate=90] () {$(8,4)$}; 
        % return pipes
        \draw[ultra thick, color=\cpassive, -latex] (31) -- (32);
        \draw[ultra thick, color=\cpassive, -latex] (32) -- (33);
        \draw[ultra thick, color=\cpassive, -latex] (33) -- (34);
        \draw[ultra thick, color=\cpassive, -latex] (31) -- (21);
        \draw[ultra thick, color=\cpassive, -latex] (22) -- (32);
        \draw[ultra thick, color=\cpassive, -latex] (23) -- (33);
        \draw[ultra thick, color=\cpassive, -latex] (34) -- (24);
        \node at (0.5*\X, -2*\Y+\dt)  [] () {$(13,14) $};
        \node at (1.5*\X, -2*\Y+\dt)  [] () {$(14,15) $};
        \node at (2.5*\X, -2*\Y+\dt) [] () {$(15,16)$};
        \node at (-\dt,-1.5*\Y) [rotate=90] () {$(13,9) $};
        \node at (\X-\dt,-1.5*\Y)  [rotate=90] () {$(10,14)$};
        \node at (2*\X-\dt,-1.5*\Y)  [rotate=90] () {$(11,15)$}; 
        \node at (3*\X-\dt,-1.5*\Y) [rotate=90] () {$(16,12)$}; 
        % active edges:
        \draw[ultra thick, color=\cslack, -latex] (21) -- (11);
        \draw[ultra thick, color=\cactive, -latex] (12) -- (22);
        \draw[ultra thick, color=\cactive, -latex] (13) -- (23);
        \draw[ultra thick, color=\cactive, -latex] (24) -- (14);
        \node at (-\dt,-0.5*\Y) [rotate=90] () {$(9,5) $};
        \node at (\X-\dt,-0.5*\Y)  [rotate=90] () {$(6,10)$};
        \node at (2*\X-\dt,-0.5*\Y)  [rotate=90] () {$(7,11)$}; 
        \node at (3*\X-\dt,-0.5*\Y) [rotate=90] () {$(12,8)$}; 
    \end{tikzpicture}
}}
        \caption{Representation as directed graph}
    \end{subfigure}
    \caption{A \gls{DH} grid (top) is modeled as a directed graph (bottom). The edges' orientation is arbitrary for passive edges, i.e., pipes (blue). For active edges (green), i.e., consumers or generators, and the slack edge (red), i.e., one heat source, it is chosen such that $\mf{ij} \geq 0$ holds in all situations.} 
    \label{fig:grid_repr}
\end{figure}
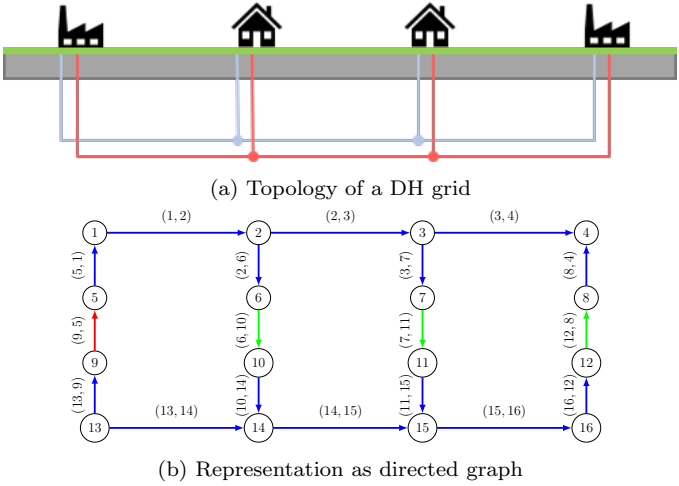

\begin{figure}
    \centering
    \scalebox{0.8}{\resizebox{\linewidth}{!}{%
    \begin{tikzpicture}
        % Nodes and Edge: 
        \node at ( 0, 3) [circle, draw] (01) {$\pr{i}$, $\temp{i}$};
        \node at (10, 3) [circle, draw] (02) {$\pr{j}$, $\temp{j}$};
        \draw[ultra thick, -latex] (01) -- (02);
        % text on top: 
        \node[] at ( 0, 4) {Node $i \in \Nodes$};
        \node[] at (10, 4) {Node $j \in \Nodes$};
        \node[] at ( 5, 4) {Edge $(i,j) \in \Edges$};
        % top mf: 
        \node at (4,3.2) (tl) {}; 
        \node at (6,3.2) (tr) {};
        \draw[-latex] (tl) -- (tr);
        \node at (5, 3.4) (t) {$\mf{ij} \geq 0$};
        % top Tend / Tstart
        \node[anchor=west] at (0.5, 3.3) {$\tempstart{ij}$};
        \node[anchor=west] at (8.2, 3.3) {$\tempend{ij}$};
        % bottom mf: 
        \node at (4,2.8) (bl) {}; 
        \node at (6,2.8) (br) {};
        \draw[-latex] (br) -- (bl);
        \node at (5, 2.6) (b) {$\mf{ij} \leq 0$};
        % top Tend 
        \node[anchor=west] at (0.5, 2.7) {$\tempend{ij}$};
        \node[anchor=west] at (8.2, 2.7) {$\tempstart{ij}$};
        
    \end{tikzpicture}
}}
    \caption{The state variables in steady-state models of DH grids include node-bound variables such as fluid temperatures $\temp{i}$ and pressures $\pr{i}$. The sign of the edge-bound mass flows $\mf{ij}$ depends on the orientation with respect to the edge orientation (indicated by the thick arrow). The localization of the temperatures at the inlet $\tempstart{}$ and outlet $\tempend{}$ of the pipe depends on the sign of $\mf{ij}$ (for positive mass flows see above the thick arrow, otherwise below).}
    \label{fig:notation}
\end{figure}
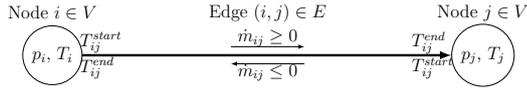

We model the \gls{DH} grids using the well-studied steady-state formulation \cite{ tol2021development, fang2014state, sun2019nonlinear}. 
That is, we neglect time delays and assume the heating fluid, typically hot water, to be incompressible and to have constant fluid properties within the relevant temperature ranges \cite{BotSte21}. 
The heating grid's toplogy is modeled as a directed graph $\Grid = (\Nodes, \Edges)$ with nodes $\Nodes$ and edges $\Edges \subseteq \Nodes \times \Nodes$.
As shown in \fig{grid_repr}, we distinguish between three types of edges:
passive edges $\EDGp$ represent pipes whose heat loss and pressure loss are determined through the grid equations. 
Active edges $\EDGa$ represent heat consumers and suppliers who exchange a specified amount of heat with the grid. 
One slack edge $\EDGsl$ in the grid, representing a heat power plant, provides any remaining balancing power and defines the pressure levels.
As visualized in \fig{notation}, we denote for each node $i \in \Nodes$ the heating fluid's temperature by $\temp{i}$ and its pressure by $\pr{i}$.
For an edge $(i, j) \in \Edges$, $\mf{ij}$ denotes the mass flow rate from $i$ to $j$. 
Symmetry implies that $\mf{ij} = - \mf{ji}$ and we define that positive values indicate that the mass flow direction aligns with the orientation of the edge. 
$\tempstart{ij}$ and $\tempend{ij}$ denote the fluid's temperature before and after traversing edge $(i,j) \in \Edges$. 
The heat exchange with the external world over edge $(i,j)\in\Edges$ is denoted by $\power{ij}$, where positive values denote heat consumption. 
For each node $i$, we denote the set of neighbours in the grid as $\Neighbours{i}$. 
%The notation is visualised in \fig{notation}.

Passive edges are characterised by hydraulic and thermal equations. The hydraulic system describes the interaction between mass flows and pressures. 
It considers the conservation of mass at each node, 
\begin{align}
    &\sum_{j\in\Neighbours{i}} \mf{ij} = 0, & \forall i \in \Nodes \label{eq:mconv}, 
\end{align}
and the pressure drop over a pipe, 
\begin{align}
        &\pr{i} - \pr{j} = k_{ij} \mf{ij} \abs{\mf{ij}}, & \forall (i,j) \in \EDGp, \label{eq:pipeP} 
\end{align}
where $\pressurdrop{ij}$ is a pipe-specific parameter. 
The thermal system describes the flow-dependent temperatures. It consists of the conservation of energy at each node, 
\begin{align}
    &\temp{i} = \frac{\sum_{j\in \Neighbours{i}, \mf{ji} > 0}\mf{ji} \tempend{ji}}{\sum_{j\in \Neighbours{i}, \mf{ji} > 0}\mf{ji}}, &\forall  i \in \Nodes,  \label{eq:tmix}    
\end{align}
and the temperature drop over pipes, 
\begin{subequations}
\begin{align}
    \tempend{ij} = \left(\tempstart{ij} - \tempa\right) \text{exp} \left(- a_{ij} / \mf{ij}\right) + \tempa, \nonumber \\  \forall (i,j) \in \EDGp,  \\
    \tempstart{ij} = \left\{
        \begin{array}{ll}
            \temp{i} \quad \text{if} \quad \mf{ij} \geq 0, \\
            \temp{j} \quad \text{if} \quad \mf{jj} < 0, \\
        \end{array} \right. \nonumber \\ \forall (i,j) \in \EDGp.
\end{align} \label{eq:pipeT}
\end{subequations} 

Here, $a_{ij}$ summarizes several pipe properties related to its length and the specific heat loss to the surroundings. $\tempa$ denotes the ambient temperature. We assume $\tempa$ to be equal for all pipes.
%We define the equation dynamically, depending on the flow direction, to avoid redundant variables in our system. 

Active edges represent heat consumers or suppliers. They summarise multiple physical elements into a single edge, such as valves, pumps, and heat exchangers. 
The behaviour of an active edge is determined by a local controller, who adapts the mass flow through the edge such that a predetermined heat power $\power{ij}$ is exchanged with the grid and a specific feed-in temperature $\tfi{ij}$ is met at the outlet of the edge, 
\begin{align}
    &\tempend{ij} = \tfi{ij} & \forall (i,j) \in \EDGa, \label{eq:activeTend}\\
    &\power{ij} = \mf{ij} \cp \left(\temp{i}-\tempend{ij}\right) & \forall (i,j) \in \EDGa. \label{eq:activePower}
\end{align}
Here, $\cp$ is the specific heat capacity of the heating fluid.
We define $\power{ij} \geq 0$ for heat demands and $\power{ij} \leq 0$ for heat supplies and choose the orientation of the edge $(i,j) \in \EDGa$ such that $\mf{ij} \geq 0$. 
Pressure changes over the active edges are not modelled explicitly, as local controllers adjust them as required to achieve the desired mass flow. 

The modelling equations above determine the pressures in the grid only up to a constant offset for the supply and return sides, respectively. These pressure levels are controlled by central pumps to ensure a sufficient pressure drop at each consumer and to avoid evaporation in grids with superheated water. 
Additionally, in steady-state, the total heat supply has to match the total consumption and the losses in the grid. Both requirements are ensured by including one slack generator $(i,j) \in \EDGsl$ which is modelled as
\begin{align}
    & \tempend{ij} = \tfi{ij}, & \forall (i,j) \in \EDGsl, \label{eq:Tslack}\\ 
    & \pr{i} = \prset{i}, \pr{j} = \prset{j}, & \forall (i,j) \in \EDGsl. \label{eq:pslack}
\end{align}
Again, we definine that $\mf{ij} \geq 0$ for $(i,j) \in \EDGsl$. 
In the remainder of this paper, we use fixed values for $\prset{i}$, $\prset{j}$, and $\tempa$ and do not explicitly mention the dependency of all results on these values. 
%we assume the pressure levels to be at $\pr{i} = 3.0\,bar$ and $\pr{j} = 6.5\,bar$ for $(i,j) \in \EDGsl$. Changing these values leads to an equal offset for all pressure values.
%Similarly, we set $\tempa=10 \C$ for all edges. We refrain from explicitly mentioning the dependency on this choice in the modelling equations to improve clarity. 

\bigskip
Equations \eqref{eq:mconv} - \eqref{eq:pslack} form a system of non-linear heat grid equations which can be compactly denoted as 
\begin{align}
    \statequations{\state, \vpower, \vtfi} = \mathbf{0}  \label{eq:stateEQ}
\end{align}
with a suitably defined non-linear operator $\statequations{\cdot}$ using vectorial quantities.
These are the active edges' heat power $\vpower = [\power{ij}]_{(i,j) \in \EDGa}$ and feed-in temperatures $\vtfi = [\tfi{ij}]_{(i,j) \in \EDGa \cup \EDGsl}$ as well as the grid state variables $\state = [\vtemp, \vmf, \vpr, \vtempend]$ consisting of
the grid temperatures $\vtemp = [\temp{i}]_{i \in \Nodes}$, the mass flows $\vmf = [\mf{ij}]_{(i,j) \in \Edges}$, the pressures $\vpr = [\pr{i}]_{i \in \Nodes}$, and the end of line temperatures $\vtempend = [\tempend{ij}]_{(i,j) \in \Edges}$.
%which comprise the grid state $\state = [\vtemp, \vmf, \vpr, \vtempend]$.
% and $\vprset = [\prset{i}, \prset{j} \forall (i,j) \in \EDGsl]$. 
The implicit function theorem \cite{hoehre_math_fuer_ingeneure} then states that, at least locally around every triplet $(\vpower, \vtfi, \state)$ that fulfils \eqref{eq:stateEQ}, there exists a well-defined mapping $\modelDS{\cdot}$ from the heat powers and feed-in temperatures to the corresponding network state, such that
\begin{align} 
    \state = \modelDS{\vpower, \vtfi}. \label{eq:statemapping}
\end{align}
This mapping has no closed form but can be evaluated by solving the heat grid equations \eqref{eq:stateEQ}, i.e., performing \gls{TPF} computations.
%It can also be approximated with learning-based approaches. 
%In the following chapter, we first discuss two established approaches for solving the grid equations. We then present our novel approach to generate a large number of training samples for which \eqref{eq:stateEQ} is satisfied as required for the training of learning-based models. 

\section{\gls{TPF} Computations}\label{sec:thermalComp}

\subsection{Classical Approaches}\label{subsec:classical_solver}
The literature discusses two main approaches for evaluating \eqref{eq:statemapping}, either separating the model equations into submodels which are solved alternatingly, the so-called \glsfirst{DC} \cite{LIU20161238}, or using nonlinear root-finding algorithms on \eqref{eq:stateEQ} \cite{tol2021development}. 
We shortly present both approaches and a combined method, which will serve as the benchmark for our experiments. 

\bigskip
The idea behind the \gls{DC} is to alternatingly solve the power equations \eqref{eq:activePower}, the hydraulic equations \eqref{eq:mconv} and \eqref{eq:pipeP}, and the thermal equations \eqref{eq:tmix},  \eqref{eq:pipeT}, \eqref{eq:activeTend} and \eqref{eq:Tslack}. The steps are summarised in \alg{decomp}. 
The algorithm is initialised by assigning initial temperature values to all nodes, i.e., $\vtemp = \vtinit$. 
The first step is then to calculate the mass flows at active edges $\vmfactive = [\mf{ij}]_{(i,j) \in \EDGa}$ via \eqref{eq:activeTend} and \eqref{eq:activePower}. 
Second, the hydraulic system equations \eqref{eq:mconv} and \eqref{eq:pipeP} are used to calculate all remaining mass flows in the grid. 
Given all mass flows $\vmf$, \eqref{eq:tmix} and \eqref{eq:pipeT} allow to calculate temperature losses over all pipes and the temperatures at all nodes, starting from the points where the temperature is fixed by \eqref{eq:activeTend} and \eqref{eq:Tslack}. The results of this third step allow us to update the initially assumed temperatures, including the temperature values of the inlet nodes for active edges used in \eqref{eq:activePower}. 
We then go back to the first step and calculate updated mass flows, based on the new temperatures. 
These three steps are repeated until the results converge. 
\apdx{hydraulic} and \apdx{thermal} describe in detail how graph-based algorithms can be used to efficiently solve the hydraulic and thermal systems. 

\begin{algorithm}[h!]
  \caption{Decomposed Algorithm}
  \label{alg:decomp}
  \SetAlgoLined\DontPrintSemicolon
\SetKwInOut{Parameter}{Parameter}
\KwIn{active edges' power values $\vpower$, \newline
  active edges' feed in temperature $\vtfi$} 
\Parameter{desired precision $\precisionDA$}
\KwOut{grid state $\state$}
  
\SetKwFunction{algo}{DC-algorithm}
\SetKwFunction{proc}{do\_decomposed\_step}
\SetKwProg{myalg}{Algorithm}{}{}  
\myalg{\algo{$\vpower, \vtfi, \precisionDA$}}{
  $\vtemp \gets \vtinit$\;
  $\SELoss \gets \infty$\;
  \While{$\Psi \geq \epsilon$}
  {
    $\state \gets \proc (\vpower, \vtfi, \vtemp)$\;
    $\SELoss \gets \Norm{\statequations{\state, \vpower}}^2$\;
    update $\vtemp$ from $\state$
  }
}
\SetKwProg{myproc}{Procedure}{}{}
\myproc{\proc{$\vpower, \vtfi, \vtemp$}}{
  $\vmfactive \gets $ solve power equations \eqref{eq:activeTend}, \eqref{eq:activePower}\;
  $\vmf, \vpr \gets$ solve hydraulic equations \eqref{eq:mconv}, \eqref{eq:pipeP} using $\vmfactive$\;
  $\vtemp, \vtempend \gets$ solve thermal equations \eqref{eq:tmix}, \eqref{eq:pipeT}, \eqref{eq:activeTend}, \eqref{eq:Tslack}\;
  $\state \gets [\vtemp, \vmf, \vpr, \vtempend]$\;
  \nl \KwRet {$\state$}\;}

\end{algorithm}

\bigskip
% \subsubsection*{Newton-Raphson Algorithm}
The \glsfirst{NR} is a root-finding algorithm in multiple dimensions. We use a modified version with an adaptive step size to solve the heat grid equations \eqref{eq:stateEQ}. 
The algorithm is initialised by assigning initial values to all states, i.e., $\state = \stateinit$. 
In each iteration, we evaluate the right-hand side of the heat grid equations and the local Jacobian matrix $\Jacobian= \left[\dfrac{\partial \statequations{\state, \vpower, \vtfi}}{\partial \state}\right]$. 
% , i.e., 
% \begin{align}
%     \RhsNR &= \statequations{\state, \vpower, \vtfi} , \label{eq:rhsSE} \\
%     \Jacobian &= \left[\dfrac{\partial \statequations{\state, \vpower, \vtfi}}{\partial \state}\right] . \label{eq:jacobian}
% \end{align}
The descent direction $\descdir$ is then obtained by solving $\Jacobian \descdir = - \statequations{\state, \vpower, \vtfi} $.
% \begin{align}
%     \Jacobian \descdir = - \RhsNR .
% \end{align}
We use a backtracking line search to adjust the stepsize $\stepsize$, i.e., we start with $\stepsize=1$ and multiply it by 0.5 until \\ $
\Norm{\statequations{\state + \stepsize \descdir, \vpower}} \leq \left(1 - \decparam \stepsize \right) \Norm{ \statequations{\state, \vpower}}
$,
% \begin{align}
%     \Norm{\statequations{\state + \stepsize \descdir, \vpower}} \leq \Norm{\statequations{\state, \vpower, \vtfi} + \decparam \stepsize (\Jacobian \descdir)}, \label{eq:armijo}
% \end{align}
where \\ $\decparam = 10^{-4}$ is a hyperparameter influencing the descent rate. The termination condition for the backtracking step is similar to the Armijo-Goldstein condition for sufficient descent \cite{armijo1966minimization}.

% The algorithm is outlined in \alg{NR}
% \begin{algorithm}
%   \caption{Newton Raphson algorithm}
%   \label{alg:NR}
%   \input{02_algorithms/alg_NR}
% \end{algorithm}

\bigskip
% \subsubsection*{Combined Algorithm} \label{subsec:combined}
The two algorithms described above show complementary convergence behaviour. 
The DC algorithm is robust concerning its initial condition; however, the gains in each step diminish close to the desired solution. 
On the other hand, the \gls{NR} algorithm is highly dependent on a good initial guess but shows fast convergence near the solution. 
This can be exploited by using a combined approach to reduce calculation times by starting with \gls{DC} until a point close to the desired solution is found, and the gain per step diminishes, then switching to \gls{NR}, using the last result of the \gls{DC} as the starting point. 
This combined algorithm was used previously in \cite{bott2023deep, BotSte21} and \cite{MatBotReh21} to solve the heat grid equations, and serves as the baseline for our experiments. 

\subsection{Learning-Based Approach}\label{subsec:LBmodels}
Learning-based approaches use a parametric model $\modelNN{\cdot}$ with parameters $\modelWeight$ to approximate the implicitly defined state mapping, i.e., \\ ${\modelNN{\vpower, \vtfi} \approx \modelDS{\vpower, \vtfi} = \state}$, for all relevant $\vpower, \vtfi$.
Evaluating these learned models can be multiple orders of magnitude faster compared to the classic \gls{TPF} computations, as no iterative solution algorithm is required.  
Building a learned model typically consists of three steps: 
\begin{itemize}
    \item[1.] Define the model structure $\modelNN{\cdot}$ with parameters $\modelWeight$. 
    \item[2.] Compute the training data, $\Bigl\{\left(\vpower_{i}, \vtfi_{i}\right), \state_{i}\Bigr\}_{i=1}^{\Ntrain}$ which cover the relevant input space and satisfy $\state_{i} = \modelDS{\vpower_{i}, \vtfi_{i}}$. 
    \item[3.] Train the model, i.e. minimise the loss regarding $\modelWeight$,
    \begin{align}
        \modelWeightOpt = \argmin{\modelWeight} \Loss{\modelWeight}.
        %= \argmin{\modelWeight} \Loss{\modelNN{\cdot}, \cdot} . 
        \label{eq:DnnLoss}
    \end{align}
    A typical loss is the squared error averaged over the training samples, i.e., \\ $\Loss{\modelWeight} = \frac{1}{N} \Sum{i=1}{N} \Norm{\modelNN{\vpower_{i}, \vtfi_{i}} - \state_{i}}^2$.
\end{itemize}

The literature discusses different parametric models and selecting the right model can have a large impact on the performance \cite{murphy2012machine}.
Apart from generic feed-forward \gls{DNN} structures, graph neural networks could be a promising model class. These were used by \cite{boussaid2023evaluation} for thermal grids and perform very well for electric power systems \cite{donon2019graph, donon2020neural}. 
However, we focus in this paper on the model-building process, not on optimising the performance for any specific \gls{DH} grid type or application.
Therefore, we use a simple \gls{DNN} as the learning-based model and refrain from an exhaustive model or hyperparameter optimisation. 
%A further promising model class in this context are graph neural networks, which were used by \cite{boussaid2023evaluation} for thermal grids and perform very well for electric power systems \cite{donon2019graph, donon2020neural}. 
Adjusting the weights in the third step is also done with standard optimisation algorithms \cite{zeiler2012adadelta, kingma2014adam}. 

Especially for large \gls{DH} grids, the second step of the model-building process is computationally very costly.
In this paper, we therefore propose a novel, importance-sampling-based algorithm, which allows the faster generation of training data, without losing sample accuracy. 

\section{Efficient Training Data Generation}\label{sec:IS}
A default approach to generate the training data set is to define a distribution $\vdist{\vpower, \vtfi}$ over the relevant input space, draw samples from this distribution and solve the \gls{TPF} problem \eqref{eq:statemapping} for each sample, using the classic algorithm described in Section~\ref{subsec:classical_solver}. 
However, this procedure is computationally costly due to the iterative nature of the \gls{DC} and \gls{NR} algorithms. 
We therefore propose a novel, \gls{IS} for training set generation. It consists of three parts as outlined below. Moreover, the algorithm is summarised in \alg{IS} and depicted in \fig{algos}.

\subsection{Proxy Distribution}\label{subsec:proxy_dist}
First, we define a proxy distribution $\vdist{\vmfactive, \vtfi}$ over the space of mass flows at active edges $\vmfactive$ and the feed in temperatures $\vtfi$. 
The choice of the proxy distribution is arbitrary as long as its support includes the support of the original distribution, i.e., $\supp{\vdist{\vpower, \vtfi}} \subseteq \supp{\vdist{\vmfactive, \vtfi}}$ \cite{robert1999monte}. 

\bigskip
In the following, we discuss the process of obtaining such a proxy distribution for our specific test setting. However, a similar line of thought can be applied to other distributions $\vdist{\vpower, \vtfi}$ as well. 
We model the distributions over the power and feed-in temperatures independently, i.e., \\${\vdist{\vpower, \vtfi} = \vdist{\vpower} \vdist{\vtfi}}$. 
The heat powers $\vpower$ are described as a truncated normal distribution $\ZTN{\vpowermean, \vpowercov}$ with mean power $\vpowermean$ and covariance matrix $\vpowercov$. 
The truncation restricts heat demands to positive values and heat supplies to negative values. 
The feed-in temperatures $\tfi{}$ are modelled via uniform distributions with minimum value $\vtfimin$ and maximum value $\vtfimax$. 

We define a proxy distribution over the mass flows only, and keep the feed-in temperature distribution unchanged.
Our mass flow proxy distribution is a zero-truncated normal distribution $\ZTN{\vmfactivemean, \vmfactivecov}$ with mean $\vmfactivemean$ and covariance matrix $\vmfactivecov$.
We calculate $\vmfactivemean$ by solving the heat grid equations for mean power values $\vpowermean$ and mean feed-in temperatures and extracting the active edges' mass flows from the resulting state, i.e. 
\begin{align}
    \vmfactivemean \in \state_{\mu} = \modelDS{\vpowermean, ({\vtfimin + \vtfimax})/{2}} .
\end{align}

To calculate the covariance matrix, we first approximate the standard deviation in each dimension by solving the heat grid equations for power values which equal the mean power values plus one standard deviation each, i.e., 
\begin{align}
\vmfactivemean + \vmfactivestd \in \state_{\mu+\sigma} = \modelDS{\vpowermean + \vpowerstd, ({\vtfimin + \vtfimax})/{2}}, 
\end{align}
where $\vmfactivestd = [\sigma_{\vmf}^j]_{j\in\EDGa}$ and $\vpowerstd = [\sigma_{\vpower}^j]_{j\in\EDGa}$ are vectors containing all standard deviations for the power and mass flow distribution respectively. 
By assuming, that the normalised cross-correlation between two power values matches the one between the corresponding mass flow values, we then calculate the entries of $\vmfactivecov$ as 
\begin{align}
    \vmfactivecov^{i,j} = \frac{\sigma_{\vmf}^i \sigma_{\vmf}^j}{\sigma_{\vpower}^i \sigma_{\vpower}^j} \vpowercov^{i,j} .
\end{align}

For the proxy distribution to be valid, it has to hold that $\supp{\vdist{\vpower, \vtfi}} \subseteq \supp{\vdist{\vmfactive, \vtfi}}$.
In other words, every sample with a non-zero probability under the original distribution must have a non-zero probability under the proxy distribution. 
Since the proxy distribution is a zero-truncated normal distribution, only samples with negative values for at least one entry in $\vmfactive$ have a zero probability. 
According to \eqref{eq:activePower}, such a negative mass flow would lead to negative power values at demands or positive power values at supplies, which both have a probability of zero under the original distribution. 

\subsection{Sample Generation}
To generate training samples, we draw \\ $\{\vmfactive_{i}, \vtfi_{i}\}_{i=1}^\Ntrain$ pairs from the proxy distribution. 
For each sample $(\vmfactive_{i}, \vtfi_{i})$, we calculate the corresponding grid state $\state_i$ by separately solving the hydraulic equations \eqref{eq:mconv} and \eqref{eq:pipeP} and the thermal equations \eqref{eq:tmix} - \eqref{eq:activeTend} and \eqref{eq:Tslack}. 
Lastly, we calculate the corresponding heat powers via \eqref{eq:activePower}, which results in the desired training data $\Bigl\{\left(\vpower_{i}, \vtfi_{i}\right), \state_{i}\Bigr\}_{i=1}^{\Ntrain}$. 

\begin{figure}[h!]
    \centering
    \includegraphics[width=0.9\linewidth]{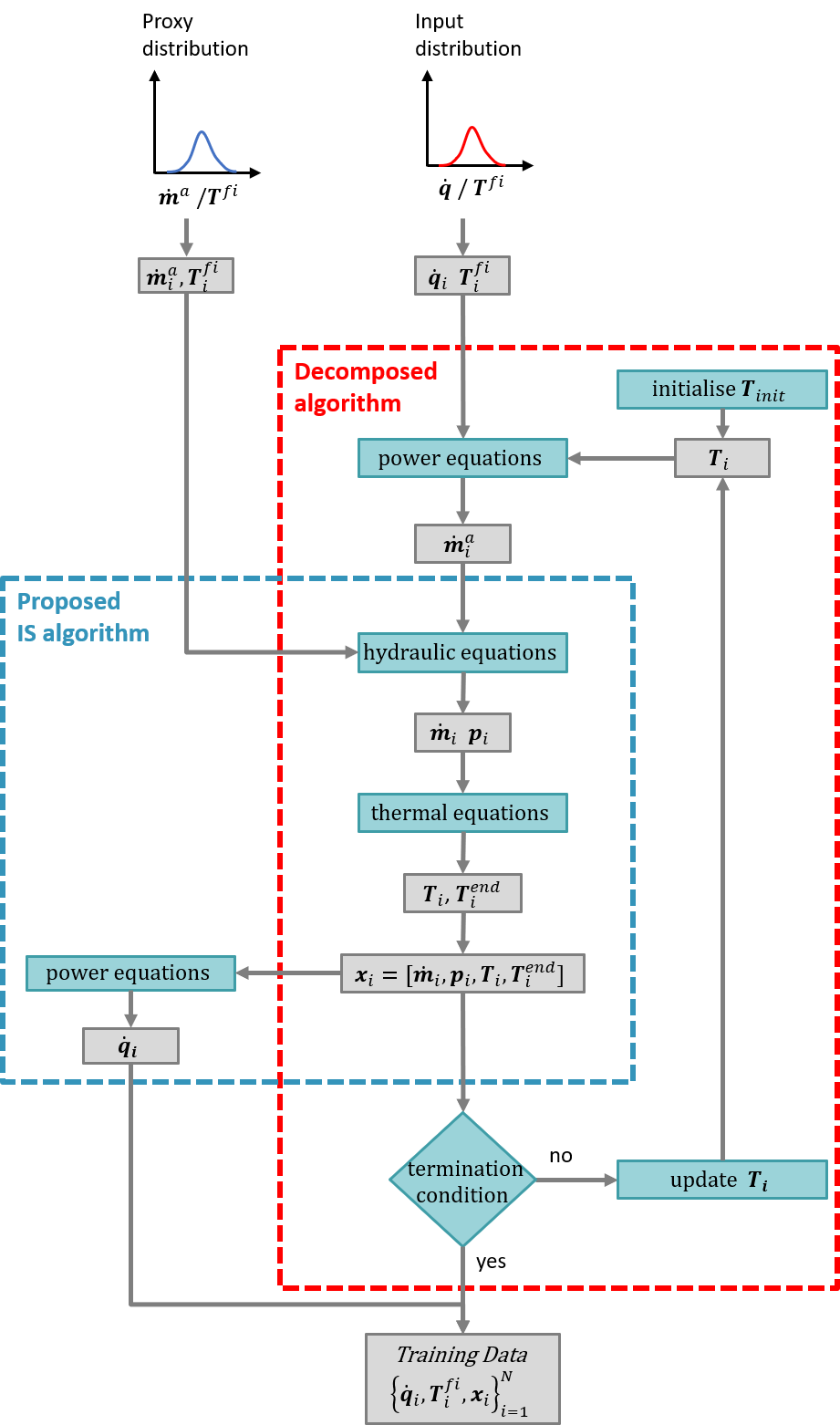}
    \caption{The default approach to generate training data is drawing $(\vpower, \vtfi)$ samples from the input distribution and solving the \gls{TPF}, e.g. by using the classic decomposed algorithm. 
    By defining a proxy distribution over $(\vmfactive, \vtfi)$ and rearranging the computational steps of the \gls{DC} algorithm, we omit the iterations and solve each training sample in a single pass.}
    \label{fig:algos}
\end{figure}

The calculation steps for each training sample match the steps of a single iteration of the \gls{DC} algorithm in a changed order. However, in our algorithm $\vpower_{i}$ becomes an output of the algorithm rather than an input, which allows us to determine the grid state in a single pass, as opposed to the iterative nature of the \gls{DC} and \gls{NR} algorithms. 
\fig{algos} illustrates the connection between our proposed algorithm and the \gls{DC} algorithm. 

\subsection{Sample Weighting}
The samples generated this way will not exactly represent the initial distribution, since they were drawn from a proxy distribution. This can be compensated by assigning a sample weight 
\begin{align}
    \weightSample{i} = \frac{\prob{\vpower_{i}, \vtfi_{i}}}{\prob{\vmfactive_{i}, \vtfi_{i}}}, \label{eq:sample_weight}
\end{align}
to each sample \cite{murphy2012machine}, where $\prob{\cdot}$ denotes the value of the corresponding probability density function, or by importance sampling resampling \cite{rubin1987calculation}. 

% These samples are then used in a second step to train a \gls{DNN}, minimising a sample-based loss function. 

\begin{algorithm}[h!]
  \caption{\gls{IS} algorithm to generate training data}
  \label{alg:IS}
  \SetAlgoLined\DontPrintSemicolon
\SetKwInOut{Parameter}{Parameter}
\KwIn{input distribution $\vdist{\vpower, \vtfi}$}
\Parameter{number of desired samples $\Ntrain$}
\KwOut{training samples $\{\left(\vpower_{i}, \vtfi_{i}\right) \state_{i}\}_{i=1}^{\Ntrain}$}

\SetKwProg{myalg}{Algorithm}{}{}
  $\vdist{\vmfactive, \vtfi} \gets$ proposal distribution for $\vdist{\vpower, \vtfi}$\;
  \For{$i=1, i\leq N, i++$}
  {
    draw sample $(\vmfactive_i, \vtfi_i) \gets \vdist{\vmfactive, \vtfi}$ \;
    $\vmf_{i}, \vpr_{i} \gets$ solve hydraulic equations \eqref{eq:mconv}, \eqref{eq:pipeP} using $\vmfactive$\;
    $\vtemp_{i}, \vtempend_{i} \gets$ solve thermal equations \eqref{eq:tmix}, \eqref{eq:pipeT}, \eqref{eq:activeTend}\;
    $\state_{i} \gets [\vtemp_{i}, \vmf_{i}, \vpr_{i}, \vtempend_{i}]$\;
    $\vpower_{i} \gets $ solve power equations \eqref{eq:activeTend}, \eqref{eq:activePower}\;
    optional: $\weightSample{i} \gets$ weight $\frac{\prob{\vpower_{i}, \vtfi_{i}}}{\prob{\vmfactive_{i}, \vtfi_{i}}}$\;
  }
  \nl \KwRet $\{\left(\vpower_{i}, \vtfi_{i}\right) \state_{i}\}_{i=1}^{\Ntrain}$
\end{algorithm}

\section{Experimental Evaluation}\label{sec:exp}
We demonstrate our proposed algorithm using grid layouts which feature typical aspects of 4\Th generation district heating grids such as multiple feed-ins and cyclic grid structures. 

\subsection{DH Grid Test Instances and Setup}\label{subsec:expsetup}
\begin{figure}
    \centering
    \begin{subfigure}[b]{0.4\textwidth}
        \centering
        \includegraphics[scale=0.6]{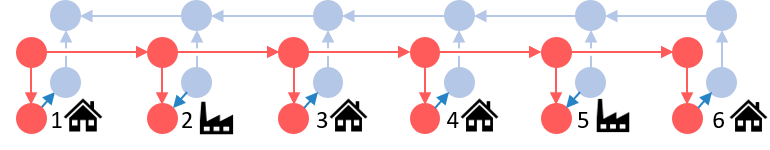}
        \caption{$\Ladder{6}{2,5}$}
        \label{fig:ladder625}
    \end{subfigure}
    \\
    \begin{subfigure}[b]{0.4\textwidth}
        \centering
        \includegraphics[scale=0.6]{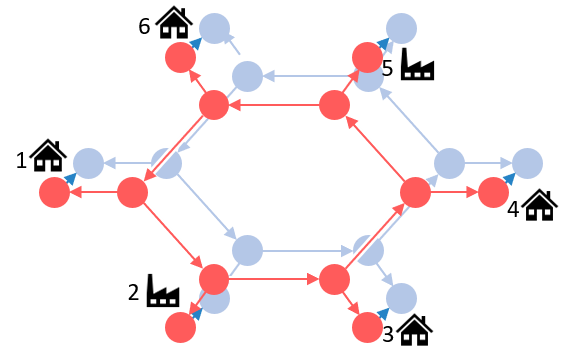}
        \caption{$\Cycle{6}{2,5}$}
        \label{fig:cycle625}
    \end{subfigure}
    \caption{In the experiments, we consider two structurally different grid layouts with varying sizes and different numbers of heat supplies.}
    \label{fig:GenerigGrids}
\end{figure}

We consider two generic grid types as shown in \fig{GenerigGrids}. 
In so-called \textit{Ladder} grids, the heat demands and supplies are lined up alongside one strand, whereas \textit{Cycle} grids consist of a pipe loop to which the active components are connected. 
We denote these grids as $\Ladder{na}{sp}$ and $\Cycle{na}{sp}$ respectively, where $na$ denotes the number of active components, i.e., demands and supplies in the grid, and $sp$ specifies the position of heat supplies. 
For example, $\Ladder{6}{2,5}$ refers to a grid with a line structure, which has heat supplies on positions 2 and 5 and heat demands on positions 1, 3, 4 and 6 as shown in \fig{ladder625}. 

% We model the distributions over the power and feed in temperatures independently, i.e., $\vdist{\vpower, \vtfi} = \vdist{\vpower} \vdist{\vtfi}$. 
% The heat powers are described as a truncated Gaussian distribution $\ZTN{\vpowermean, \vpowercov}$ with mean power $\vpowermean$ and covariance matrix $\vpowercov$. 
% The truncation restricts heat demands to positive values and heat supplies to negative values. 
% Feed-in temperatures are modelled as uniform distributions with minimum values $\vtfimin$ and maximum values $\vtfimax$. 

As discussed in Section~\ref{subsec:proxy_dist}, we model the distributions over the heat powers and feed-in temperatures as $\vdist{\vpower, \vtfi} = \vdist{\vpower} \vdist{\vtfi}$ where \\ $\vdist{\vpower} = \ZTN{\vpowermean, \vpowercov}$.  
The heat demands are assumed to have an equal mean consumption. The correlation between demands decreases with their respective distance. 
The heat supplies have an equal mean power value each and are uncorrelated. 
The temperatures are modelled as uniform distributions between minimum values $\vtfimin$ and maximum values $\vtfimax$.
% As shown in \ref{subsec:proxy_dist}, the proxy distribution over the active edges' mass flows is a zero truncated normal distribution as well. 
Details regarding the numerical values including $\tempa$ and $\prset{}$ can be found in \apdx{numvalgrids}. 

The reported computation times were achieved on a local compute server with a 22 Core Intel i6 processor and 78 GB of RAM. 
The \gls{DNN} training using the sample-based loss, was done with an NVIDIA GeForce GTX 1080. 
We did not use the GPU for the other calculations, as this led to significantly higher computation times. 
No special means were taken to parallelise the code besides optimisations done by the TensorFlow backend \cite{tensorflow2015-whitepaper}. 

The code for our experiments will be published on GitHub on paper acceptance. 

\subsection{Efficient Training Data Generation} \label{subsec:expIS}
We now compare our proposed \gls{IS} approach for training data generation against the default approach with the combined \gls{TPF} algorithm presented in Section~\ref{subsec:classical_solver}. 
All experiments are executed 5 times each. All values are reported as averages over these 5 runs plus/minus one standard deviation. 
We assess the quality of our proxy distribution by calculating the effective sample rate \cite{liu1996metropolized}
\begin{align}
    \effSR = \frac{1}{1 + \Var{\weightsampleNorm{}}},     \label{eq:effSR}
\end{align}
where $\Var{\weightsampleNorm{}}$ is the variance of the normalised sample weights according to \eqref{eq:sample_weight}.

\begin{figure}
    \centering
    \includegraphics[width=\linewidth]{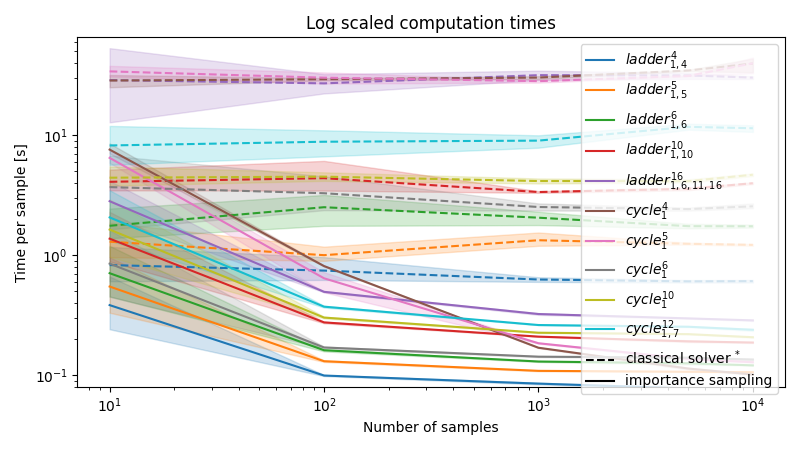}
    \caption{Comparison between the calculation times using the default approach (dashed lines) and the proposed approach (solid lines) to calculate different numbers of training samples. 
    The impact of the additional overhead for calculating the proxy distribution introduced with the proposed approach diminishes for large sample sizes.}
    \label{fig:sampleTimes}
\end{figure}

\bigskip
\fig{sampleTimes} shows the computation times per training sample for different \gls{DH} grid topologies and different numbers of generated samples. 
\tab{is_red} repeats the numerical results exemplarily for the two test sets $\Cycle{12}{1,7}$ and $\Ladder{16}{1,6,11,16}$. 
Further numerical results can be found in \tab{IS_res_p1} and \tab{IS_res_p2} in the Appendix. 
Our proposed algorithm introduces additional computational overhead to generate the proxy distribution $\vdist{\vmf}$ but computing each sample is much faster. 
As the overhead is independent of the number of training samples, it becomes less relevant for larger sample sizes. 
The effective sample rate increases with the number of drawn samples and exceeds 99\% for all grids when sampling over 1000 samples. 
\begin{table*}[]
    \centering
    \caption{Comparison between the computation times for both algorithms (time reduction factor) and the effective sample rate for two exemplary grids.}
    \renewcommand{\arraystretch}{1.2}
\begin{tabular}{llr@{}lr@{}l}
& \# samples&  \multicolumn{2}{c}{time reduction factor} & \multicolumn{2}{c}{effective sample rate} \\ 
\hline
\parbox[t]{2mm}{\multirow{5}{*}{\rotatebox[origin=c]{90}{ $\Ladder{16}{1.6.11.16}$}} }
& 10      & 12.3 $\pm$   & 11.6 &  91.2\% $\pm$        &  0.5\%            \\
& 100     & 54.3 $\pm$   & 11.0 &  99.3\% $\pm$        &  0.2\%            \\
& 1000    &   97.3 $\pm$ & 10.9 &  $\geq$ 99.9\% $\pm$ &  0.02\%           \\
& 5000    &  105 $\pm$   & 6.92 &  $\geq$ 99.9\% $\pm$ & $\leq$ 0.01\%     \\
& 10 000  & 104 $\pm$  & 4.04 &  $\geq$ 99.9\% $\pm$ & $\leq$ 0.01\%     \\
\hline
\parbox[t]{2mm}{\multirow{5}{*}{\rotatebox[origin=c]{90}{$\Cycle{12}{1.7}$}}}
& 10     & 5.40 $\pm$   & 3.27  &   92.6\% $\pm$        &  2.2\%            \\ 
& 100    & 23.7 $\pm$   & 6.96  &   99.2\% $\pm$        &  0.2\%            \\ 
& 1000   &   34.2 $\pm$ & 4.51  &   $\geq$ 99.9\% $\pm$ &  0.03\%           \\
& 5000   & 46.1 $\pm$   & 3.18  &   $\geq$ 99.9\% $\pm$ & $\leq$ 0.01\%     \\ 
& 10 000 & 47.5 $\pm$   & 3.10  &   $\geq$ 99.9\% $\pm$ & $\leq$ 0.01\%    

\end{tabular}

    \label{tab:is_red}
\end{table*}

\begin{figure}
    \centering
    \includegraphics[width=\linewidth]{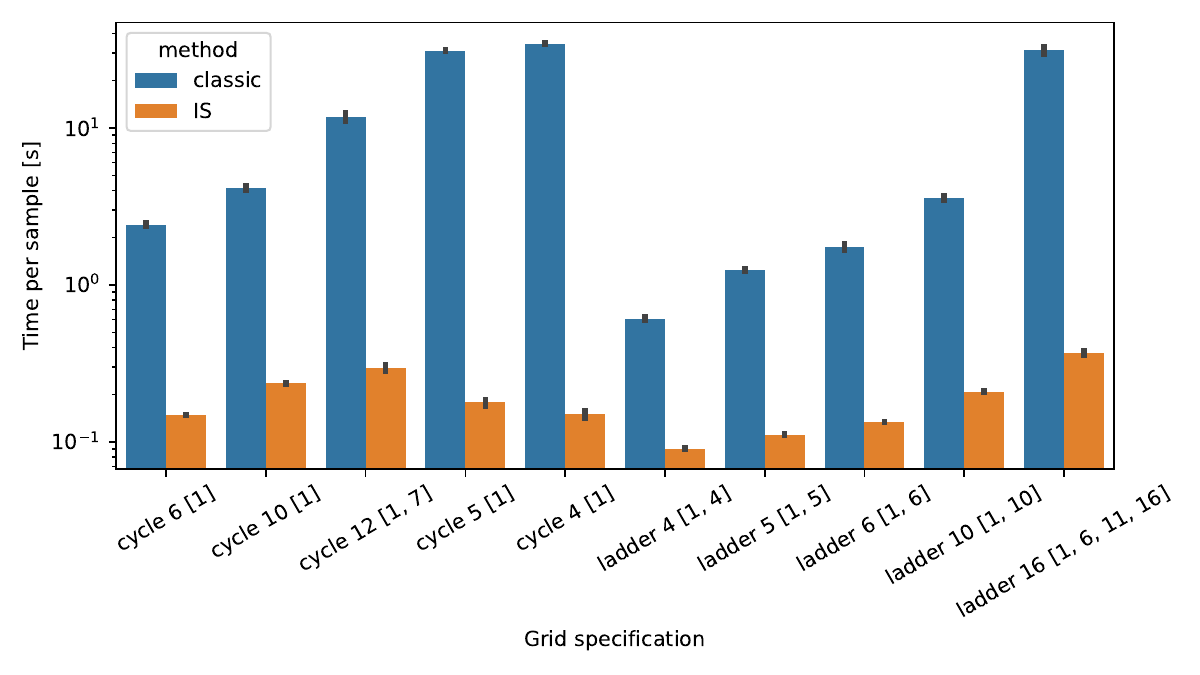}
    \caption{The sampling times for 10000 samples vary depending on the grid topology. Our proposed algorithm seems to scale better for complex grids. However, it is not directly evident what makes a grid complex.}
    \label{fig:complexity}
\end{figure}

\fig{complexity} shows a concise comparison of the computation times per training sample when drawing 10000 samples for each grid layout. 
The lowest relative time gain is achieved in the $\Ladder{4}{1,4}$ test case with a factor of 7.4, the fastest network to solve with the classic algorithm. 
The $\Cycle{4}{1}$ case, on the other hand, reaches the highest computation times for the classic algorithm and has the largest relative time gains when comparing both approaches. The proposed algorithm is 387 times faster for this test case. 
The largest sampling times for the proposed algorithm are reached for the $\Ladder{16}{1,6,11,16}$ test case, in which it is about 100 times faster than the default approach. 

It is difficult to draw a conclusion, on how the algorithms scale for larger grids, as calculation times do not necessarily correlate to grid sizes, especially for the cycle grids. 
When comparing the fastest and slowest computation times for cycle grids for both algorithms, the differences are a factor of 15.5 for the classic approach and 2.5 for the proposed algorithm. 
For the ladder grids, the factors are 49.3 and 3.7, respectively. 
Thus, we assume that our proposed algorithm is less prone to increase computation times with increasing grid complexity than the default approach. 

\subsection{\gls{DNN} Training}\label{subsec:expDNN}
We now investigate, how well the generated training data perform for training a learning-based \gls{TPF} model. 
To this end, we approximate the state mapping \eqref{eq:statemapping} with a \gls{DNN}. We use the same \gls{DNN} structure independent of the heat grid layout. 
The core of the network are four fully connected trainable layers. The first three layers have [200, 400, 400] neurons and a rectified linear unit as an activation function. The fourth layer, whose size depends on the number of state variables, has a linear activation function. 
The trainable layers are surrounded by layers with fixed weights used for data scaling, internal postprocessing, and to exclude trivial model outputs from the training, e.g., $\tempend{ij}$ if $(i,j) \in \EDGa$.  
The detailed model structure is found in \apdx{DNNstructure}.

The model then has the form
\begin{align}
    \modelNN{\vpower, \vtfi} = \statepred,
\end{align}
where $\statepred$ is the approximation for the grid state.
We compare the traditional model training approach, using a weighted squared loss function against a sample-free physics-aware loss function.

\subsubsection{Sample-Based Training}
In this common training setup, the \gls{DNN} is trained by minimising the weighted squared loss function 
\begin{align}
    \LossW{\modelWeight} = \frac{1}{N}\Sum{i=1}{N} \Sum{\kappa \in \{\temp{}, \mf{}, \pr{}, \tempend{}\}}{} \weightLoss{\kappa} \Norm{\statepred^{\kappa}_i - \state^{\kappa}_i}^2. 
\end{align}
where $\weightLoss{\kappa}$ is a dimension-dependent weight, compensating for the different orders of magnitude of the entries in $\state$ and increasing the numerical stability of the training process. Note that the selection of weights depends on the units of measurement used. 
For our experiments, we use $kg/s$ for mass flows, $\C$ for temperatures and $bar$ for pressures, which leads to 
\begin{align}
    \weightLoss{\kappa}  = 
    \begin{cases} 
        500 &\quad \text{if} \quad \kappa = \mf{} \\
         1 &\quad \text{else} .
    \end{cases}
\end{align}

For the model training, we use 50 000 training samples and the Adam \cite{kingma2014adam} optimiser with TensorFlow standard parameters and a batch size of 32. 
We use a validation set of 10 000 samples and stop the training if the validation loss no longer decreases over 50 epochs. We then use the weights with the lowest validation loss for further evaluation. 
The very high effective sample rates achieved in the data generation allow us to refrain form reweighting each individual training sample according to \eqref{eq:sample_weight} during the model training.

\subsubsection{Physics-Aware Training}\label{subsec:PAL}
In the physics-aware training setup, we define the loss function as 
\begin{align}    
    \LossPA{\modelWeight} = \frac{1}{N}  \Sum{i=1}{N} 
    \Norm{\statequations{\modelNN{\vpower_{i}, \vtfi_{i}}, \vpower_{i}, \vtfi_{i}}}^2. \label{eq:PALloss}
\end{align}
This loss directly penalises the violation of the heat grid equations \eqref{eq:stateEQ} of the \gls{DNN}'s output and is independent of the true corresponding state $\state_i$. 
Since this loss solely relies on samples of the form $\{(\vpower_{i}, \vtfi_{i})\}_{i=1}^{\Ntrain}$ which can be easily sampled from the initial distribution, this approach allows to skip the sample generation step. 

Each training epoch consists of 10 000 training samples. The samples are generated independently for each epoch.
We use the Adadelta \cite{zeiler2012adadelta} optimiser with an initial learning rate of 0.01 and a batch size of 32. 
The training is stopped if the training loss does not decrease over 20 epochs. 
We use the weights with the lowest loss for further evaluation. 
Since new training samples are generated in each epoch, we can monitor the training loss without requiring dedicated validation data. 

\subsubsection{Evaluation Metrics}
To evaluate the performance of the resulting \gls{DNN}, we compute three error scores. 
The first error score is the \gls{RMSE} of the predicted states $\statepred$. 
For the second error score, we summarise the passive heat grid equations \eqref{eq:mconv}, \eqref{eq:pipeP}, \eqref{eq:tmix} and \eqref{eq:pipeT} for all edges and nodes as $
    \stateequationsPas{\state} = \mathbf{0}  \label{eq:stateEQPassive}
$
and define a physical consistency metric as
\begin{align}
    \LossSE = \frac{1}{N} \frac{1}{2 (\abs{\Edges} + \abs{\Nodes})} \Sum{i=1}{N} \Norm{\stateequationsPas{\statepred_i}}^2  , 
\end{align}
where $\abs{\Edges}$ and $\abs{\Nodes}$ denote the cardinality of the sets of edges and nodes, respectively.
Note that due to the structure of the \gls{DNN}, the heat grid equations \eqref{eq:activeTend}, \eqref{eq:Tslack} and \eqref{eq:pslack} are always fulfilled and therefore excluded from the loss calculation. 
Third, we calculate the predicted heat powers $\vpowerpred$ by solving \eqref{eq:activePower} using $\statepred$ and calculate the \gls{RMSE} between the predicted and the true heat powers. 

The first error score evaluates the overall performance of the \gls{DNN} and is commonly used in many \gls{ML} applications.  However, it is agnostic to the task. 
The other two scores assess problem-specific properties of solving the \gls{TPF} problem. 
The physical consistency metric assesses, how close the predicted state is to fulfilling physical feasibility, whereas the last loss assesses specifically the heat powers' deviation from the true solution since these are especially relevant for many applications. 

\subsubsection{Results}
\tab{DNNresultsRed} compares the three error scores for the two training schemes for different grid layouts.
\begin{table*}[]
    \centering
    \caption{Achieved error scores for the \gls{DNN} using the sample-based loss function and the physics-aware loss function. The training with the sample-based loss reaches lower error scores for most metrics and grids. 
    Regarding calculation times, the training times (column "training") are similar for both approaches; however, the physics-aware training approach does not require calculating samples beforehand (column "data").}
    \begin{tabular}{llccccc}
\multirow{2}{*}{grid\_spec}& \multirow{2}{*}{Loss Function}& \multirow{2}{*}{\gls{RMSE} $\state$} & \multirow{2}{*}{$\LossSE$ } & \gls{RMSE} $\vpower$ & training &  data \\
                           &                         &                                      &                             & [kW]                 &   [h:mm]&   [h:mm]\\ 
\hline
\multirow{2}{*}{$\Ladder{4}{1.4}$}          & physics-aware  & 3.36 & 1.75& 65.3&  \underline {0:07}&    ---\\
                                            & sample-based   & \underline {0.54} & \underline {0.49}& \underline {2.39} &  0:48&    1:48\\
                                            \rule{0pt}{3ex}
% \hline
\multirow{2}{*}{$\Ladder{5}{1.5}$}          & physics-aware  & 2.47  & \underline {0.01}& 46.1  &   \underline {0:27}&    ---\\
                                            & sample-based & \underline {0.55} & 1.11 & \underline {2.89}&   0:34&    2:07\\
                                            \rule{0pt}{3ex}
% \hline
\multirow{2}{*}{$\Ladder{6}{1.6}$}          & physics-aware  & 2.55  & \underline {0.01} & 46.5  &  \underline {0:46}&    ---\\
                                            & sample-based & \underline {0.56} & 0.50 & \underline {2.44}&  0:51&    2:22\\
                                            \rule{0pt}{3ex}
% \hline
\multirow{2}{*}{$\Ladder{10}{1.10}$}        & physics-aware  & 2.43& 1.26 & 97.6&  0:35&    ---\\
                                            & sample-based & \underline {0.64} & \underline {0.37} & \underline {3.91} &  \underline {0:28} &    3:52\\
                                            \rule{0pt}{3ex}
% \hline
\multirow{2}{*}{$\Ladder{16}{1.6.11.16}$}   & physics-aware  & 9.89& \underline {0.13} & 107.8&  1:04&    ---\\
                                            & sample-based & \underline {1.05}& 1.37 & \underline {6.11}&  \underline {0:36}&    5:55\\
                                            \rule{0pt}{3ex}
% \hline
\multirow{2}{*}{$\Cycle{4}{1}$ }            & physics-aware  & 0.67 & \underline {0.01} & 39.3&  \underline {0:25}&    ---\\
                                            & sample-based & \underline {0.17} & 0.10 & \underline {2.44}&  0:38&    2:03\\
                                            \rule{0pt}{3ex}
% \hline
\multirow{2}{*}{$\Cycle{5}{1}$}             & physics-aware  & 3.51& 0.43 & 213.2&   \underline {0:15}&    ---\\
                                            & sample-based & \underline {0.15} & \underline {0.13} & \underline {1.14}&  0:28&    2:32\\
                                            \rule{0pt}{3ex}
% \hline
\multirow{2}{*}{$\Cycle{6}{1}$}             & physics-aware  & 4.42& 2.44& 129.3&  \underline {0:11}&    ---\\
                                            & sample-based & \underline {0.77} & \underline {0.62} & \underline {3.76}&  0:21&    2:39\\
                                            \rule{0pt}{3ex}
% \hline
\multirow{2}{*}{$\Cycle{10}{1}$}            & physics-aware  & 2.72& \underline {0.20} & 86.8&  0:33&    ---\\
                                            & sample-based & \underline {0.78} & 0.65 & \underline {4.83} &  \underline {0:22}&    4:06\\
                                            \rule{0pt}{3ex}
% \hline
\multirow{2}{*}{$\Cycle{12}{1.7}$}          & physics-aware  & 3.58 & \underline {0.14} & 203.8& 0:48&    ---\\
                                            & sample-based & \underline {0.82} & 0.84 & \underline {5.44}& \underline {0:24}& 4:46
\end{tabular}

    \label{tab:DNNresultsRed}
\end{table*}
The sample-based approach tends to reach a lower \gls{RMSE} for the state, as to be expected since this metric resembles the one minimised during the training. 
The evaluation of the physical consistency metric $\LossSE$, which is part of the loss function of the second approach, offers less clear results. 
For some tested topologies, the sample-free approach yields significantly lower errors, for other ones, the sample-based approach does. 
The \gls{RMSE} for the power values show the most significant performance deviation between both training schemes. Here, the sample-based approach reaches much better scores, even though the power deviation is part of \eqref{eq:PALloss}. 

\bigskip
The training time per batch is higher for the physics-aware loss. However, this is counteracted by the lower number of training samples and fewer required number of epochs compared to the sample-based training scheme. In total, we obtain comparable training times for both approaches. 
The training data generation takes significantly longer than the model training. 
However, by using the proposed approach to generate training samples, computation times for this step were reduced significantly. 
Considering, that the model building has to be done only once, the lower error scores in the relevant metrics 1 and 3 for the sample-based training scheme outweigh the additional computation times and favour this approach over the physics-aware training scheme for most practical applications. 
% However, using the algorithm proposed in this paper, the computation times for this step could be brought down to a few hours, which 
% Even though the data generation takes significantly longer than the model training, the lower error scores for the sample-based training scheme clearly favour this approach over the physics-aware training scheme. 

\section{Conclusion}\label{sec:concl}
Learning-based approaches to \gls{TPF} computations can perform very well and can be very fast. 
Their training, however, traditionally requires a large number of training samples, which are computationally costly to generate. 
We use a simple \gls{DNN} as a learning-based model and compare a traditional training scheme, minimising a weighted squared loss between the predicted and the true solution, with a sample-free physics-aware training scheme. 
The sample-based training scheme reaches a significantly lower \gls{RMSE} for the predicted state and heat powers. Regarding the physical consistency metric, neither approach clearly outperforms the other. 

A major share of the computational burden of the sample-based training approach lies within the generation of training data. 
We propose a novel \gls{IS} algorithm, to produce input-output-tuples for typical input values of the \gls{TPF} computation more efficiently.
To this end, we define a proxy distribution over the space of active edges' mass flows, draw samples from it, and calculate the corresponding grid state and heat powers. 
In our experiments, the proposed algorithm reduces the computation times per generated sample by one to two orders of magnitude, compared to the default approach of drawing heat powers and solving the \gls{TPF}.
For demand values following a truncated normal distribution, we reach high effective sample rates of over $99.9\%$ in all test setups. 

Our algorithm for efficiently generating consistent consistent tuples $\Bigl\{\left(\vpower_{i}, \vtfi_{i}\right), \state_{i}\Bigr\}_{i=1}^{\Ntrain}$ is motivated by the generation of training samples for the learning-based \gls{TPF} approaches. 
% We use the samples generated this way to train our learning-based model for the \gls{TPF}. 
However, the same approach could be used also for other tasks which require a large number of grid state samples, e.g., in the area of district heating network design or for sample-based probabilistic state estimation. 
% The required large number of training samples can be generated efficiently using the proposed algorithm. 

% While we use our algorithm to generate training data for learning-based \gls{TPF} models, the approach is not limited to this application. Further research could apply the algorithm to other tasks, which require solving the \gls{TPF} for randomised inputs, such as sample-based state estimation. 

\section{Acknowledgements}
This research was funded by the German Federal Ministry for Economic Affairs and Climate Action (BMWK) under project number 03EN3012A.

% \printbibliography
\bibliographystyle{ieeetr}
\bibliography{01_formalities/main}
\appendix
\section{Solving the hydraulic model}\label{app:hydraulic}
Equation \eqref{eq:mconv} can be rewritten as 
\begin{align}
    \mathbf{I} \vmf = \mathbf{0}
\end{align}
where $\mathbf{I}$ denotes the incidence matrix of the graph $\Grid$. The heating grid is divided between pipes transporting the hot water from the heat supplies to the demands and the pipes returning the cooled water back to the heat supplies. We will call the former the supply side of the network and the latter the return side. We can therefore write 
\begin{align}
    \mathbf{I} \vmf = \mathbf{I}_s \vmf + \mathbf{I}_r \vmf + \mathbf{I}_a \vmf = \mathbf{0}, \label{eq:mfconvapp}
\end{align}
where $\mathbf{I}_s$, $\mathbf{I}_r$ and $\mathbf{I}_a$ are the incidence matrices belonging to the supply side, the return side or the active edges respectively. 
The supply side and the return side are only connected via active edges. 
This allows to split up \eqref{eq:mfconvapp} in two systems of equations containing either $\mathbf{I}_s$ and $\mathbf{I}_a$ or $\mathbf{I}_r$ and $\mathbf{I}_a$. 
Substituting $\mathbf{I}_a \vmf = \pm \vmfactive$ leads to a system of linear equations to be solved. 

If the grid contains a loop, the solution for these equations is not unique. Including the additional condition, that the sum of all pressure losses in the loop equals zero leads to a system with one quadratic equation. 
Once all mass flows are known, it is trivial to calculate all pressures based on \eqref{eq:pipeP} and \eqref{eq:pslack}. 

\section{Solving the thermal System}\label{app:thermal}
If all mass flows are known, the temperatures are solved by propagating from active edges through the grid. 
Initially, all nodes and all edges in $\EDGp$ are marked as unsolved and all edges in $\EDGa$ and $\EDGsl$ as solved, as their end-of-line temperature $\tempend{}$ is fixed via \eqref{eq:activeTend} and \eqref{eq:Tslack}.
In each iteration, we check for every node, whether all edges from whom water flows into the node are marked as solved. 
If this is the case, the temperature $\temp{}$ of this node can be calculated via \eqref{eq:tmix}. The node is then marked as solved and the end-of-line temperature $\tempend{}$ of all edges, whose water flow origins in this node, can be solved via \eqref{eq:pipeT}. These edges are then marked as solved as well. 
This process is repeated until all nodes are marked as solved. 
Note, that the computational burden can be reduced significantly by only checking nodes which are connected to edges which were newly marked as solved in the previous iteration.

\section{Numerical Values for Test Instances} \label{app:numvalgrids}
The heat power distribution and the feed-in temperature distribution are modelled independently. 
The heat powers follow a truncated Gaussian distribution $\ZTN{\vpowermean, \vpowercov}$. 
For each demand $i$, we assume the mean heat consumption to be $\vpowermean^{i} = 200\,kW$ and the standard deviation to be 20\% of the consumption, i.e. $\vpowercov^{i,i} = 160\, kW^2$. 
The mean power value for heat supplies is chosen equally for all demands, such that the total mean production matches the total mean demand. The standard deviation is set to be 20\% of the mean power as well. 
We assume no correlation between heat supplies or between heat supplies and heat demands. 
In between heat demands, we assume the cross-correlation of the power values to decrease, with the distance $\ldem{i,j}$ between two demands $i$ and $j$ according to 
\begin{align}
    \demcor{ij} &= \exp\left(-\corrfactor \frac{\ldem{i,j}}{\max_{ij} \ldem{i,j}}\right), 
\end{align}
where $\corrfactor$ is a free correlation-strength-factor. 
We measure $\ldem{i,j}$ by counting demands, i.e. $\ldem{i,j}=1$ for neighbours and $\ldem{i,j}=2$ if one demand or supply is between $i$ and $j$. We set $\corrfactor=5$ for our experiments. 
% \begin{align}
%     d &= 
%     \begin{cases}
%         \abs{i - j} &if gridtype = \Ladder{}{} \\
%         \min(\abs{i-j}, (N - \abs{i-j})) & else
%     \end{cases}
%     \label{eq:corr}
% \end{align}

The feed-in temperatures for heat demands are fixed at $55\,\C$, for heat supplies the minimal value is $90\,\C$ and the maximum is $130\,\C$. 
We use $\tempa = 10\C$ and set the pressure values to $\prset{i} = 3.5\, bar$ and $\prset{j} = 6.5 bar$ with $ (i,j) \in \EDGsl$.

\section{Result tables}
\begin{landscape}
    \begin{table}[]
        \centering
        \caption{Numerical results for sampling 10 - 1000 samples. All numbers are reported as mean +/- 1 std, averaged over 5 runs}
        \begin{tabular}{llr@{}lr@{}lr@{}lr@{}lr@{}l}
\# samples               & grid specification        & \multicolumn{2}{c}{sampling time NR [s]} & \multicolumn{2}{c}{setup time IS [s]} & \multicolumn{2}{c}{Sampling time IS [s]} & \multicolumn{2}{c}{effective sample rate} &  \multicolumn{2}{c}{time reduction factor} \\ 
\hline
\multirow{10}{*}{10}     & $\Ladder{4}{1.4}$        &   8.3  $\pm$ &  3.3     &  2.9 $\pm$ &  2.74	  &  1.0  $\pm$ &  0.07     & 92.8\% $\pm$ &  2.2\%      &   2.66 $\pm$ &  0.68         \\
                         & $\Ladder{5}{1.5}$        &  13.0  $\pm$ &  5.6     &  4.3 $\pm$ &  4.04    &  1.2  $\pm$ &  0.06     & 91.3\% $\pm$ &  0.7\%      &   2.92 $\pm$ &  0.59         \\
                         & $\Ladder{6}{1.6}$        &  17.6  $\pm$ &  6.8     &  5.6 $\pm$ &  4.83    &  1.5  $\pm$ &  0.08     & 93.6\% $\pm$ &  1.6\%      &   3.04 $\pm$ &  0.63         \\
                         & $\Ladder{10}{1.10}$      &  40.8  $\pm$ & 10.7     & 11.5 $\pm$ &  9.04    &  2.2  $\pm$ &  0.15     & 92.2\% $\pm$ &  2.4\%      &   3.75 $\pm$ &  0.95         \\
                         & $\Ladder{16}{1.6.11.16}$ &  20.0  $\pm$ & 23.4     & 24.8 $\pm$ &  13.3    &  3.4  $\pm$ &  0.33     & 91.2\% $\pm$ &  0.5\%      &   12.3 $\pm$ &  11.6         \\
                         & $\Cycle{4}{1}$           &   284  $\pm$ & 38.3     & 73.9 $\pm$ &  8.12    &  1.8  $\pm$ &  1.24     & 98.4\% $\pm$ &  0.9\%      &   4.26 $\pm$ &  0.62         \\
                         & $\Cycle{5}{1}$           &   338  $\pm$ & 42.0     & 62.4 $\pm$ &  8.55    &  2.3  $\pm$ &  1.65     & 98.4\% $\pm$ &  1.2\%      &   6.21 $\pm$ &  0.46         \\
                         & $\Cycle{6}{1}$           &  36.9  $\pm$ & 33.0     &  5.9 $\pm$ &  5.36    &  2.6  $\pm$ &  2.21     & 92.7\% $\pm$ &  1.9\%      &   6.82 $\pm$ &  7.26         \\
                         & $\Cycle{10}{1}$          &  44.2  $\pm$ & 10.5     & 12.2 $\pm$ &  9.08    &  4.1  $\pm$ &  3.54     & 92.4\% $\pm$ &  1.8\%      &   3.70 $\pm$ &  1.07         \\
                         & $\Cycle{12}{1.7}$        &  81.9  $\pm$ & 36.1     & 16.1 $\pm$ &  10.4    &  4.5  $\pm$ &  3.68     & 92.6\% $\pm$ &  2.2\%      &   5.40 $\pm$ &  3.27         \\ \\
\multirow{10}{*}{100}    & $\Ladder{4}{1.4}$        &  74.6  $\pm$ & 21.9     &  1.4 $\pm$ &  0.02	 &  8.6 $\pm$ &   0.11      & 99.2\% $\pm$ &  0.2\%      &   7.51 $\pm$ &  2.20         \\ 
                         & $\Ladder{5}{1.5}$        &  100   $\pm$ & 17.4     &  2.1 $\pm$ &  0.05    & 11.0 $\pm$ &   0.22     & 99.4\% $\pm$ &  0.3\%      &   7.70 $\pm$ &  1.40         \\
                         & $\Ladder{6}{1.6}$        &  251   $\pm$ & 87.1     &  3.1 $\pm$ &  0.21    & 13.2 $\pm$ &   0.24     & 99.2\% $\pm$ &  0.2\%      &   15.5 $\pm$ &  5.36         \\
                         & $\Ladder{10}{1.10}$      &  437   $\pm$ & 171      &  7.0 $\pm$ &  0.33    & 20.6 $\pm$ &   0.28     & 99.4\% $\pm$ &  0.2\%      &   15.9 $\pm$ &  6.06         \\
                         & $\Ladder{16}{1.6.11.16}$ &  286   $\pm$ & 242      & 18.6 $\pm$ &  0.38    & 31.1 $\pm$ &   0.22     & 99.3\% $\pm$ &  0.2\%      &   54.3 $\pm$ &  11.0         \\
                         & $\Cycle{4}{1}$           & 2912   $\pm$ & 71.4     & 71.1 $\pm$ &  0.80    & 10.3 $\pm$ &   0.12     & $\geq$ 99.9\% $\pm$ &  0.04\%     &   36.2 $\pm$ &  0.77         \\
                         & $\Cycle{5}{1}$           & 3001   $\pm$ & 392      & 51.1 $\pm$ &  15.8    & 13.3 $\pm$ &   0.64     & 99.8\% $\pm$ &  0.2\%      &   50.2 $\pm$ &  12.4         \\
                         & $\Cycle{6}{1}$           &  328   $\pm$ & 123      &  2.9 $\pm$ &  0.17    & 14.2 $\pm$ &   0.21     & 99.3\% $\pm$ &  0.3\%      &   19.4 $\pm$ &  7.49         \\
                         & $\Cycle{10}{1}$          &  451   $\pm$ & 41.9     &  8.0 $\pm$ &  0.36    & 22.3 $\pm$ &   0.35     & 99.4\% $\pm$ &  0.2\%      &   15.0 $\pm$ &  1.54         \\
                         & $\Cycle{12}{1.7}$        &  880   $\pm$ & 256      & 11.8 $\pm$ &  0.32    & 25.5 $\pm$ &   0.63     & 99.2\% $\pm$ &  0.2\%      &   23.7 $\pm$ &  6.96         \\ \\
\multirow{10}{*}{1000}   & $\Ladder{4}{1.4}$        &  628   $\pm$ & 32.0     &  1.4 $\pm$ &  0.05	  & 84.1 $\pm$ &   1.61	 & $\geq$ 99.9\% $\pm$ &  0.03\%   &   7.35 $\pm$ &  0.33         \\ 
                         & $\Ladder{5}{1.5}$        & 1334   $\pm$ & 203      &  2.3 $\pm$ &  0.12    & 107  $\pm$ &   0.97     & $\geq$ 99.9\% $\pm$ &  0.03\%   &   12.2 $\pm$ &  1.92         \\
                         & $\Ladder{6}{1.6}$        & 2054   $\pm$ & 310      &  3.2 $\pm$ &  0.16    & 127  $\pm$ &   1.08     & $\geq$ 99.9\% $\pm$ &  0.03\%   &   15.7 $\pm$ &  2.27         \\
                         & $\Ladder{10}{1.10}$      & 3358   $\pm$ & 63.1     &  7.3 $\pm$ &  0.16    & 203  $\pm$ &   0.93     & $\geq$ 99.9\% $\pm$ &  0.02\%   &   16.0 $\pm$ &  0.30         \\
                         & $\Ladder{16}{1.6.11.16}$ & 2692   $\pm$ & 583      & 18.2 $\pm$ &  0.60    & 306  $\pm$ &   5.53     & $\geq$ 99.9\% $\pm$ &  0.02\%   &   97.3 $\pm$ &  10.9         \\
                         & $\Cycle{4}{1}$           &30070   $\pm$ & 974      & 69.1 $\pm$ &  0.85    & 101  $\pm$ &   0.66     & $\geq$ 99.9\% $\pm$ &  $\leq$ 0.01\% &    177 $\pm$ &  6.56         \\
                         & $\Cycle{5}{1}$           &28090   $\pm$ & 781      & 56.4 $\pm$ &  1.22    & 129  $\pm$ &   1.24     & $\geq$ 99.9\% $\pm$ &  $\leq$ 0.01\% &    152 $\pm$ &  4.63         \\
                         & $\Cycle{6}{1}$           & 2520   $\pm$ & 199      &  3.0 $\pm$ &  0.12    & 140  $\pm$ &   1.30     & $\geq$ 99.9\% $\pm$ &  0.03\%   &   17.6 $\pm$ &  1.24         \\
                         & $\Cycle{10}{1}$          & 4144   $\pm$ & 239      &  8.2 $\pm$ &  0.22    & 218  $\pm$ &   1.26     & $\geq$ 99.9\% $\pm$ &  0.03\%   &   18.3 $\pm$ &  1.02         \\
                         & $\Cycle{12}{1.7}$        & 8987   $\pm$ & 1198     & 11.8 $\pm$ &  0.19    & 251  $\pm$ &   0.43     & $\geq$ 99.9\% $\pm$ &  0.03\%   &   34.2 $\pm$ &  4.51         \\
\end{tabular}

        \label{tab:IS_res_p1}
    \end{table}
\end{landscape}
    \newpage
\begin{landscape}
    \begin{table}[]
        \centering
        \caption{Numerical results for sampling 5000 - 10 000 samples. All numbers are reported as mean +/- 1 std, averaged over 5 runs}
        \begin{tabular}{llr@{}lr@{}lr@{}lr@{}lr@{}l}
\# samples               & grid specification        & \multicolumn{2}{c}{sampling time NR [s]} & \multicolumn{2}{c}{setup time IS [s]} & \multicolumn{2}{c}{Sampling time IS [s]} & \multicolumn{2}{c}{effective sample rate} &  \multicolumn{2}{c}{time reduction factor} \\ 
\hline
\multirow{10}{*}{5000}   & $\Ladder{4}{1.4}$        &   3034 $\pm$ &  84.7     &  1.4 $\pm$ & 0.05        &  389 $\pm$ &   4.9       & $\geq$ 99.9\% $\pm$ & $\leq$ 0.01\% & 7.77 $\pm$ &  0.25         \\
                         & $\Ladder{5}{1.5}$        &   6231 $\pm$ &   129     &  2.4 $\pm$ & 0.09        &  534 $\pm$ &   1.3       & $\geq$ 99.9\% $\pm$ & $\leq$ 0.01\% & 11.6 $\pm$ &  0.24         \\
                         & $\Ladder{6}{1.6}$        &   8729 $\pm$ &   432     &  3.1 $\pm$ & 0.11        &  628 $\pm$ &    12       & $\geq$ 99.9\% $\pm$ & $\leq$ 0.01\% & 13.8 $\pm$ &  0.69         \\
                         & $\Ladder{10}{1.10}$      &  17855 $\pm$ &   577     &  6.7 $\pm$ & 0.26        &  954 $\pm$ &    19       & $\geq$ 99.9\% $\pm$ & $\leq$ 0.01\% & 18.6 $\pm$ &  0.42         \\
                         & $\Ladder{16}{1.6.11.16}$ & 156561 $\pm$ & 10019     & 17.6 $\pm$ & 0.16        & 1474 $\pm$ &   9.8       & $\geq$ 99.9\% $\pm$ & $\leq$ 0.01\% &  105 $\pm$ &  6.92         \\
                         & $\Cycle{4}{1}$           & 172228 $\pm$ &  2566     & 68.8 $\pm$ & 1.32        &  502 $\pm$ &    10       & $\geq$ 99.9\% $\pm$ & $\leq$ 0.01\% &  302 $\pm$ &  6.91         \\
                         & $\Cycle{5}{1}$           & 155028 $\pm$ &  1662     & 47.4 $\pm$ & 14.5        &  640 $\pm$ &   5.6       & $\geq$ 99.9\% $\pm$ & $\leq$ 0.01\% &  226 $\pm$ &  6.41         \\
                         & $\Cycle{6}{1}$           &  12107 $\pm$ &   432     &  3.0 $\pm$ & 0.09        &  700 $\pm$ &   2.2       & $\geq$ 99.9\% $\pm$ & $\leq$ 0.01\% & 17.2 $\pm$ &  0.63         \\
                         & $\Cycle{10}{1}$          &  20676 $\pm$ &   759     &  8.1 $\pm$ & 0.10        & 1095 $\pm$ &   4.3       & $\geq$ 99.9\% $\pm$ & $\leq$ 0.01\% & 18.7 $\pm$ &  0.70         \\
                         & $\Cycle{12}{1.7}$        &  58533 $\pm$ &  3861     & 11.8 $\pm$ & 0.22        & 1260 $\pm$ &   7.6       & $\geq$ 99.9\% $\pm$ & $\leq$ 0.01\% & 46.1 $\pm$ &  3.18         \\ \\
\multirow{10}{*}{10 000} & $\Ladder{4}{1.4}$        &   6094 $\pm$ &   163     &  1.3 $\pm$ & 0.19        &  767 $\pm$ &   6.4       & $\geq$ 99.9\% $\pm$ & $\leq$ 0.01\% & 7.93 $\pm$ &  0.26         \\ 
                         & $\Ladder{5}{1.5}$        &  12202 $\pm$ &   245     &  2.3 $\pm$ & 0.10        & 1062 $\pm$ &   6.2       & $\geq$ 99.9\% $\pm$ & $\leq$ 0.01\% & 11.5 $\pm$ &  0.21         \\
                         & $\Ladder{6}{1.6}$        &  17428 $\pm$ &   584     &  3.0 $\pm$ & 0.07        & 1211 $\pm$ &   2.6       & $\geq$ 99.9\% $\pm$ & $\leq$ 0.01\% & 14.4 $\pm$ &  0.49         \\
                         & $\Ladder{10}{1.10}$      &  39766 $\pm$ &    795    &  6.4 $\pm$ & 0.31        & 1873 $\pm$ &    15       & $\geq$ 99.9\% $\pm$ & $\leq$ 0.01\% & 21.2 $\pm$ &  0.42         \\
                         & $\Ladder{16}{1.6.11.16}$ & 300512 $\pm$ &  8665     & 16.3 $\pm$ & 0.61        & 2856 $\pm$ &  33         & $\geq$ 99.9\% $\pm$ & $\leq$ 0.01\% & 104.7 $\pm$ & 4.04
     \\
                         & $\Cycle{4}{1}$           & 393638 $\pm$ &  59107    & 65.0 $\pm$ & 0.51        &  953 $\pm$ &   3.2       & $\geq$ 99.9\% $\pm$ & $\leq$ 0.01\% &  387 $\pm$ &  58.1         \\
                         & $\Cycle{5}{1}$           & 395247 $\pm$ &  67282    & 50.4 $\pm$ & 1.22        & 1239 $\pm$ &    11       & $\geq$ 99.9\% $\pm$ & $\leq$ 0.01\% &  306 $\pm$ &  49.8         \\
                         & $\Cycle{6}{1}$           &  25547 $\pm$ &   1495    &  2.8 $\pm$ & 0.12        & 1353 $\pm$ &    32       & $\geq$ 99.9\% $\pm$ & $\leq$ 0.01\% & 18.9 $\pm$ &  1.29         \\
                         & $\Cycle{10}{1}$          &  46786 $\pm$ &   1317    &  7.3 $\pm$ & 0.32        & 2073 $\pm$ &    23       & $\geq$ 99.9\% $\pm$ & $\leq$ 0.01\% & 22.5 $\pm$ &  0.86         \\
                         & $\Cycle{12}{1.7}$        & 113927 $\pm$ &   8253    & 10.9 $\pm$ & 0.79        & 2387 $\pm$ &    58       & $\geq$ 99.9\% $\pm$ & $\leq$ 0.01\% & 47.5 $\pm$ &  3.10   
\end{tabular}

        \label{tab:IS_res_p2}
    \end{table}
\end{landscape}

\section{Structure of the used DNN}\label{app:DNNstructure}
The trainable layers (\textit{ReLu1-3, Linear1}) are surrounded by layers with fixed weights to scale the inputs (\textit{InputScaling}) and outputs (\textit{OutputScaling}). Trivial outputs are assigned directly via the \textit{Assigment} Layer and concatenated to the output (\textit{StateOutput}). The power values are calculated based on the state prediction as the second output (PowerOutput). 
Here $n_a$ denotes the number of active grid components (heat supplies or demands) and $n_{out}$ is the number of predicted state variables. $n_{fix} = 2 n_a +2$ is the number of fixed trivial state variables. We fix $\tempend{ij}, \temp{j} \forall (i,j) \in \EDGa \cup \EDGsl$ and $\pr{i}, \pr{j} \forall (i,j) \in \EDGsl$. 
    
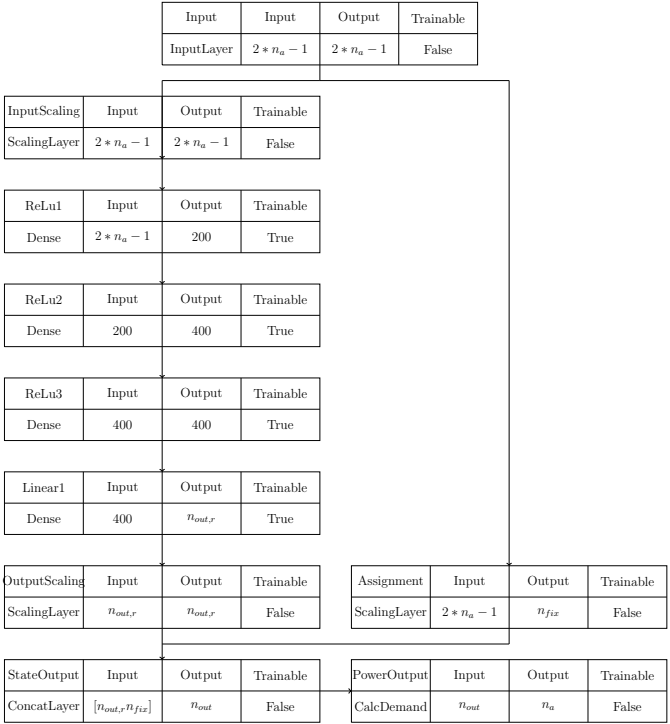
\begin{figure}
    \scalebox{1}{\tikzset{
    pics/NNnode/.style args={#1,#2,#3,#4,#5,#6}{
        code={
            % draw grid: 
            \draw (-5,-1) -- (5,-1) -- (5,1) -- (-5,1) -- (-5,-1);
            \draw (-2.5,-1) -- (-2.5,1);
            \draw (0,-1) -- (0,1);
            \draw (2.5,-1) -- (2.5,1);
            \draw (-5,0) -- (5,0);
            % layer name block
            \node[black] (name_#1) at (-3.75, 0.5){#2};
            % layer type block
            \node[black] (type_#1) at (-3.75,-0.5) {#3};
            % layer input block
            \node[black] (label_input_#1) at (-1.25,0.5) {Input};
            \node[black] (input_#1) at (-1.25,-0.5) {#4};
            % layer output block
            \node[black] (label_output_#1) at (1.25,0.5) {Output};
            \node[black] (output_#1) at (1.25,-0.5) {#5};
            % layer trainable block
            \node[black] (label_trainable_#1) at (3.75,0.5) {Trainable};
            \node[black] (trainable_#1) at (3.75,-0.5) {#6};
            % reference points for connections
            \coordinate[black] (#1_top) at (0,1) {};
            \coordinate[black] (#1_bottom) at (0,-1) {};
            \coordinate[black] (#1_left) at (-5,0) {};
            \coordinate[black] (#1_right) at (5,0) {};
        }
    }
}
\resizebox{\linewidth}{!}{%
    \begin{tikzpicture}
        \draw (5,  0) pic{NNnode={input, Input, InputLayer, $2*n_{a}-1$, $2*n_{a}-1$, False}};
        \draw (0, -3) pic{NNnode={scalingInp, InputScaling, ScalingLayer, $2*n_{a}-1$, $2*n_{a}-1$, False}};
        \draw (0, -6) pic{NNnode={Relu1, ReLu1, Dense, $2*n_{a}-1$, $200$, True}};
        \draw (0, -9) pic{NNnode={Relu2, ReLu2, Dense, $200$, $400$, True}};
        \draw (0,-12) pic{NNnode={Relu3, ReLu3, Dense, $400$, $400$, True}};
        \draw (0,-15) pic{NNnode={Linear1, Linear1, Dense, $400$, $n_{out,r}$, True}};
        \draw (0,-18) pic{NNnode={scalingOut, OutputScaling, ScalingLayer, $n_{out,r}$, $n_{out,r}$, False}};
        \draw (11,-18) pic{NNnode={assign, Assignment, ScalingLayer, $2*n_{a}-1$, $n_{fix}$, False}};
        \draw (0,-21) pic{NNnode={concat, StateOutput, ConcatLayer, $[n_{out,r} n_{fix}]$, $n_{out}$, False}};
        \draw (11,-21) pic{NNnode={power, PowerOutput, CalcDemand, $n_{out}$, $n_{a}$, False}};
    
        % draw line between inputs and next layers: 
        \coordinate (helper_inp1) at (5,-1.5);
        \coordinate (helper_inpl) at (0,-1.5);
        \coordinate (helper_inpr) at (11,-1.5);
        \draw[thick, black] (input_bottom) -- (helper_inp1);
        \draw[thick, black] (helper_inp1) -- (helper_inpl);
        \draw[thick, black] (helper_inp1) -- (helper_inpr);
        \draw[thick, black, ->] (helper_inpl) -- (scalingInp_bottom);
        \draw[thick, black, ->] (scalingInp_bottom) -- (Relu1_top);
        \draw[thick, black, ->] (Relu1_bottom) -- (Relu2_top);
        \draw[thick, black, ->] (Relu2_bottom) -- (Relu3_top);
        \draw[thick, black, ->] (Relu3_bottom) -- (Linear1_top);
        \draw[thick, black, ->] (Linear1_bottom) -- (scalingOut_top);
        \draw[thick, black, ->] (helper_inpr) -- (assign_top);
        \coordinate (helper_out_l) at (0,-19.5);
        \coordinate (helper_out_r) at (11,-19.5);
        \draw[thick, black] (scalingOut_bottom) -- (helper_out_l);
        \draw[thick, black] (assign_bottom) -- (helper_out_r);
        \draw[thick, black] (helper_out_r) -- (helper_out_l);
        \draw[thick, black, ->] (helper_out_l) -- (concat_top);
        \draw[thick, black, ->] (concat_right) -- (power_left);
    \end{tikzpicture}
}}
    \caption{Structure of the used DNN.}
    \label{fig:NN_scheme}
\end{figure}

% \begin{thebibliography}{00}
% % \bibitem{b2} Clarke, Arthur C. 2001: A Space Odyssey. New York: Roc, 1968. 297.
% \end{thebibliography}
%TC:endignore

% Word count
% \verbatiminput{\jobname.wordcount.tex}

\end{document}